\documentclass[pdflatex,sn-mathphys-num,iicol]{sn-jnl}

\usepackage{graphicx}%
\usepackage{multirow}%
\usepackage{amsmath,amssymb,amsfonts}%
\usepackage{amsthm}%
\usepackage{mathrsfs}%
\usepackage[title]{appendix}%
\usepackage{xcolor}%
\usepackage{textcomp}%
\usepackage{manyfoot}%
\usepackage{booktabs}%
\usepackage{algorithm}%
\usepackage{algorithmicx}%
\usepackage{algpseudocode}%
\usepackage{listings}%
\usepackage{xcolor}

\theoremstyle{thmstyleone}%
\theoremstyle{thmstyletwo}%

\theoremstyle{thmstylethree}%

\begin{document}

\title[Article Title]{Semantic Privacy Protection with Utility Preservation for 3D Point Clouds}







\author[1]{\fnm{Jinchang} \sur{Zhang}}

\author[1]{\fnm{Jiakai} \sur{Lin}}

\author[1]{\fnm{David} \sur{Crandall}}

\author*[1]{\fnm{Guoyu} \sur{Lu}}\email{guoyulu62@gmail.com}

\affil*[1]{\orgdiv{Department of Computer Science},
  \orgname{Indiana University},
  \orgaddress{\city{Bloomington}, \state{IN}, \postcode{47405}, \country{USA}}}


\abstract{
Point cloud data face serious semantic privacy risks during acquisition, transmission, and cross-institutional sharing. Existing methods mostly rely on geometric perturbation or destructive encryption, which can reduce the recognizability of the original class but often impair downstream usability. This paper proposes a class-transfer-based semantic encryption framework for point clouds, aiming to conceal original class information while preserving task utility and supporting authorized recovery. Specifically, we construct a unified latent space with a shared-backbone Normalizing Flow, and combine LoRA and FiLM to achieve parameter-efficient class-conditional adaptation. We further introduce diffusion-guided flow alignment to regularize the latent distribution, construct an energy-based category transition graph, and obtain an optimal class-transfer table through global matching. Then, a latent-space Neural ODE continuously evolves source-class latents into target-class latents, which are decoded into target-class point clouds through the inverse flow. We adopt attacker-oriented metrics, including New-Class Recognition Rate (NCRR), Original-Class Leakage Rate (OCLR), and Original Label Recovery Rate (OLRR), to evaluate privacy and utility. Experiments on classification and segmentation benchmarks show that the proposed method achieves controllable semantic transformation, effectively reduces original-class semantic leakage, preserves downstream learnability in the protected domain, and supports reliable authorized reconstruction.
}




\maketitle

\section{Introduction}\label{sec1}

In recent years, three-dimensional (3D) point cloud deep learning has made rapid progress in tasks such as autonomous driving, robotic perception, industrial inspection, and 3D reconstruction. With the development of 3D sensing technologies such as LiDAR, modern sensing devices are now capable of capturing high-precision point clouds of real-world environments~\cite{zhang2024embodiment,zhang2025vision}. As a result, point clouds have become a core representation for tasks including classification, segmentation, localization, and 3D reconstruction~\cite{zhang2024underground,lin20253d}.
Despite their value in 3D vision, point clouds often contain a large amount of sensitive information, such as pedestrians, vehicles, and detailed environmental structures. Once such data are uploaded to the cloud, shared with third-party platforms, or used in cross-institution collaborative training, unauthorized access, analysis, or retraining may lead to privacy leakage and exposure of commercial information~\cite{guzman2021unravelling,de2019first}. Therefore, how to reduce original semantic leakage while preserving the downstream utility of point cloud data has become a key challenge in 3D vision.
In the 2D image domain, extensive research has been conducted on privacy-preserving learning~\cite{liu2024game,liu2024stable}, including removing sensitive attributes or replacing features to protect content~\cite{barattin2023attribute,dave2022spact}, constructing adversarial examples to disrupt model learning~\cite{liu2023diffprotect,shamshad2023clip2protect}, and adopting cryptographic techniques such as differential privacy and homomorphic encryption~\cite{yonetani2017privacy}. However, due to fundamental differences in data representation, these methods are difficult to transfer directly to 3D point clouds. Point clouds are not only discrete and unordered 3D geometric samples, but their privacy risks also often arise not merely from local pixels or local attributes, but from the overall object category semantics and their spatial structure.

Existing point cloud privacy protection methods mainly reduce the recognizability of original point clouds through coordinate shuffling, noise injection, optical chaotic encryption, or hybrid-key mechanisms~\cite{liu2023privacy,li20243d,jia2019encryption}. However, such methods usually damage geometric structure, thereby weakening the usability of the protected data for downstream tasks. \cite{wang2024unlearnable} introduces an unlearnability-oriented design, making protected point clouds harder for unauthorized models to exploit. Although these methods can suppress original-class recognition to some extent, they often sacrifice the structural stability and downstream utility of the protected data, and controllable, reversible semantic-level protection for point clouds remains underexplored.
However, in many real-world scenarios, data owners do not need to “destroy point clouds so thoroughly that they become unusable,” but rather to “prevent the exposure of original sensitive semantics while still allowing the shared point clouds to be used for authorized tasks.” For example, in collaborative autonomous driving development, point cloud data may need to be shared with third-party perception model providers for continued training of detection or segmentation modules; in industrial inspection and warehouse logistics, 3D scanned data may need to be shared with collaborators for scene understanding, defect recognition, or robotic grasping; in robotic perception and 3D reconstruction, shared point clouds often still need to retain downstream learnability. If only strongly destructive geometric perturbations or local occlusion are applied, the original class may indeed become harder to recognize, but the shared data may simultaneously lose geometric interpretability and training value. Therefore, when the privacy risk comes from overall category semantics rather than local attributes, merely relying on attribute suppression, local occlusion, or destructive geometric transformations is insufficient.
Motivated by this, we study a semantic-level privacy protection problem: given an original point cloud sample, we aim to transform it into another reasonable target-category representation without exposing its original category semantics, while preserving downstream utility as much as possible and supporting authorized recovery of the original representation. In other words, our goal is not to make the data entirely unlearnable, but to reduce the risk of original-category semantic leakage while retaining the value of the data for tasks such as classification and segmentation.
Our adopts a reversible class-to-class transformation because, if what must be concealed is the overall category identity of an object, merely deleting local regions or perturbing local geometry is often insufficient to eliminate global category cues; meanwhile, unguided strong geometric destruction may effectively reduce original-class recognition, but usually cannot guarantee that the transformed sample remains a semantically coherent and learnable point cloud. In contrast, class-to-class semantic transformation moves a sample from the original class semantic manifold to another reasonable target-class semantic manifold. This both weakens the direct semantic correspondence between the shared sample and its original class, thereby reducing the risk of original semantic leakage, and preserves the structural plausibility and learnability of the sample in the protected domain.
To the best of our knowledge, this is the first semantic encryption framework for 3D point clouds that achieves semantic-level privacy protection through class-conditioned reversible transformations. Specifically, we first construct a Normalizing-Flow-based latent representation framework that maps point clouds into a shared latent space, and introduce LoRA~\cite{hu2022lora} and FiLM~\cite{perez2018film} into the Flow backbone to enable parameter-efficient multi-class modeling and lightweight adaptation. Based on this, we define a Category Transition Energy between every pair of classes and formulate category assignment as a minimum-cost matching problem to obtain an optimal category mapping table. Finally, we introduce a Neural-ODE-based continuous reversible flow in the latent space, allowing samples to migrate smoothly between class-specific manifolds and be decoded back into point cloud space through the inverse Flow, thereby generating encrypted point clouds with transformed categories. Our framework explicitly pursues three objectives simultaneously: suppressing original-category semantic leakage, preserving downstream utility in the protected domain, and supporting authorized recovery.

\begin{figure*}[t]
\begin{center}
\includegraphics[width=16cm, height=3.5cm]{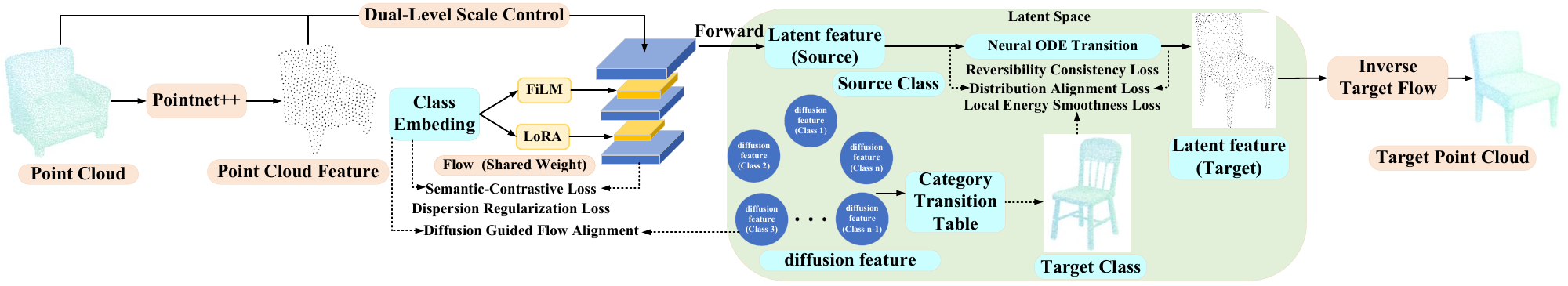}
\end{center}
\vspace{-6mm}
\caption{Overview of the framework. First, the input point cloud is encoded by PointNet++ into point cloud features, which are fed into a shared-weight Flow model together. Under the modulation of class embeddings, LoRA and FiLM jointly provide dual-level scale control, yielding the source latent representation. In the latent space, the corresponding target class is determined by the category transition table, and a neural ODE is used to continuously evolve the source latent into the target latent representation. Finally, the target latent is decoded into a target-class point cloud via the inverse target Flow, achieving semantic-level point cloud encryption and class-wise transformation.
}
\vspace{-3mm}
\label{arch}
\end{figure*}

This paper is directly relevant to the theme of the special issue, “Trustworthy Open-World Visual Recognition.” The special issue explicitly highlights privacy and security risks in visual recognition systems, and specifically welcomes research on privacy-preserving techniques and adaptation to open environments. This work studies the problem of semantic privacy protection in 3D point cloud sharing scenarios. Its core objective is to reduce the risk of original-class semantic leakage while preserving the learnability and recoverability of the protected representation, which makes it highly consistent with the privacy-preserving techniques direction emphasized by the special issue. In addition, the proposed framework further considers the incremental adaptation capability to unseen categories, which also resonates with the need for dynamic expansion in open environments. Overall, this paper provides a new semantic-level solution for trustworthy privacy protection in 3D vision.

We summarize our contributions as follows:
1. To the best of our knowledge, we propose the first semantic encryption framework for 3D point clouds, which achieves semantic-level privacy protection by applying class-conditioned reversible transformations to point cloud samples.
2. We integrate LoRA and FiLM into the Flow model to enable parameter-efficient fine-tuning with shared weights.
3. We employ a Normalizing Flow to construct the latent representation of point clouds and compute a Category Energy Graph.
4. A Neural ODE Flow is adopted to realize continuous class-to-class transformations in latent space. Our framework is showed in Fig.~\ref{arch}.

\section{Related work}

\subsection{Visual Privacy Protection}

With the rapid development of computer vision, privacy protection for 2D images and videos has been extensively investigated. Early work by Sweeney~\cite{sweeney2002k} systematically defined the notion of data privacy and highlighted potential privacy risks inherent in visual data such as images and videos.
More recently, a range of representative studies~\cite{barattin2023attribute, jiang2023dartblur} have focused on facial privacy protection, aiming to conceal identity-related attributes while preserving the usability of visual data for downstream tasks including detection, recognition, and tracking. For instance, GAN approaches~\cite{barattin2023attribute,maximov2020ciagan} synthesize faces with controlled identity characteristics. 
Adversarial methods~\cite{shamshad2023clip2protect} protect privacy by introducing carefully designed perturbations that mislead recognition systems.
Overall, prior work has convincingly demonstrated the feasibility of privacy protection in the 2D visual domain. Nevertheless, most existing methods are inherently tailored to 2D data, and privacy protection for 3D point clouds remains comparatively underexplored. Unlike images, 3D point clouds encode geometric structures, leading to fundamentally different privacy risks and protection requirements. 
Recent point cloud understanding methods have focused on jointly modeling global context and local geometry, as demonstrated by GPSFormer~\cite{wang2024gpsformer}. Dynamic acoustic field fitting~\cite{wang2025point} further improves local geometric representation for point cloud understanding. For few-shot point cloud semantic segmentation, TaylorSeg~\cite{wang2025taylor} formulates the task as a local polynomial fitting problem and captures multi-order geometric structure. DyPolySeg~\cite{wang2025dypolyseg} extends this idea through dynamic polynomial fitting for stronger few-shot segmentation performance. Reasoning Beyond Points~\cite{wang2025reasoning} further enhances few-shot 3D segmentation through a visual introspective reasoning framework beyond point-level matching.
For example,  \cite{izmailov2020semi} introduce a flow-based generative framework that perturbs point cloud geometry while retaining categorical semantics, thereby striking a trade-off between geometric distortion and semantic preservation.
\cite{liu2023privacy} proposed a point cloud privacy protection method based on optical chaotic encryption.  \cite{li20243d} further proposed a point cloud encryption method based on hybrid key and spatial maintenance. These methods mainly reduce the recognizability of the original point cloud through geometric perturbation, but do not consider the task usability of the protected point cloud after encryption. By contrast, UMT \cite{wang2024unlearnable} constructs “unlearnable” 3D point clouds through class-related transformations, making it difficult for unauthorized users to train effective models on the protected data; however, its performance in the transformed domain largely depends on a fixed transformation-label correspondence.
In contrast, our method explicitly models inter-class relationships and controlled migration processes at the semantic level. The key point is that relying only on geometric perturbation can usually reduce the recognizability of the original category, but it is difficult to simultaneously guarantee the stability and usability of the protected representation. Methods that depend on a fixed transformation-label correspondence are also often constrained by predefined transformation rules in their transformed-domain performance. In this work, we formulate the protection process as a category-level semantic redirection, so that the point cloud can suppress leakage of the original class semantics while still forming a protected representation with stable organizational structure. Therefore, compared with existing methods, the main characteristic of our approach is that it not only focuses on making the original point cloud unrecognizable, but also further emphasizes making the protected representation still learnable and usable after encryption, which makes it more suitable for application scenarios where both privacy protection and downstream task usability must be considered simultaneously.

\subsection{Normalizing Flow Models}

Normalizing Flow models constitute a class of generative models with strictly invertible architectures, which transform complex data distributions into more tractable latent distributions through bijective mappings. Representative models such as RealNVP~\cite{dinh2016density} and Glow~\cite{kingma2018glow} employ affine coupling layers and invertible $1{\times}1$ convolutions, respectively, enabling efficient modeling of high-dimensional distributions while maintaining exact invertibility. Owing to these properties, normalizing flows have been widely applied to image generation, speech modeling, and three-dimensional point cloud generation, including PointFlow~\cite{yang2019pointflow}.  
To further improve controllability, Conditional Normalizing Flows ~\cite{ardizzone2020conditional} incorporate auxiliary conditions into the flow mapping, thereby supporting conditional generation and multimodal reversible transformations.
With the advent of continuous-time dynamical systems, flow-based models have been reformulated within continuous frameworks, such as  Neural Ordinary Differential Equations~\cite{xia2021heavy}, which learn smooth vector fields connecting noise and data distributions and often lead to more stable training and convergence. Building on these developments,  Diffusion Flow~\cite{liu2022flow} integrates diffusion noise scheduling with flow models, combining robustness with computational efficiency.  
Despite these advances, most existing flow-based approaches are primarily designed for generation tasks, and their potential for semantic-level encryption of point clouds remains largely unexplored. In this work, we therefore formulate the flow transformation as an invertible encryption process that enables category-level conversion of point clouds, thereby providing effective semantic privacy protection.

\section{Method}

\subsection{Problem Definition }

This paper does not merely focus on point cloud encryption, but rather studies a semantic privacy protection problem that better aligns with practical point cloud sharing needs. Given an original point cloud sample, our goal is to transform it into another plausible target-class representation without exposing its original class semantics, while preserving as much utility as possible for downstream tasks and supporting recovery of the original representation under authorized conditions. In other words, our objective is not to make the data completely unlearnable, but to reduce the risk of original class semantic leakage while retaining its value for classification, segmentation, and subsequent modeling tasks.
This problem is motivated by clear real-world scenarios. In applications such as autonomous driving, industrial inspection, warehouse logistics, robotic perception, and 3D scanning, point cloud data often need to be shared across different organizations, platforms, or algorithmic modules. However, these point clouds usually contain sensitive class-level semantic information, such as specific equipment, goods, vehicle types, facility categories, or environmental semantics. If the original point clouds are shared directly, the receiving party may recover these sensitive semantics through training or inference, thereby causing privacy leakage or exposure of commercially sensitive information. Therefore, the practical need is not to destroy the data completely, but to ensure that the shared data remain usable for downstream tasks without revealing the original sensitive semantics.
Under this setting, we assume that the data owner only releases semantically transformed point clouds and does not disclose the original point clouds. An attacker can access the transformed samples and attempt to recover or infer their original class semantics. Accordingly, the “privacy” considered in this paper means that the original class semantics should not be directly identifiable or effectively recoverable from the shared point clouds.
Category transfer can mitigate privacy risk because it controllably moves a sample from the semantic manifold of its original class to that of another target class. After transformation, the sample no longer maintains a direct semantic correspondence with the original class, which reduces the attacker’s ability to identify or recover the original semantics. At the same time, because the sample remains within a plausible target-class semantic space, its structural interpretability and task learnability can be preserved as much as possible.

\subsection{Conditional Flow for Reversible Mapping}

This section presents a point cloud encryption and reversible transformation framework built upon Conditional Normalizing Flows. The proposed model jointly takes point clouds and their associated features as input, and learns a class-conditioned reversible mapping of point cloud distributions. By enforcing class-conditional constraints in the latent space, the framework achieves stable intra-class representations while ensuring clear separability across different classes, enabling effective encrypted transformations.

\textbf{Conditional Flow Model:}
Given an input point cloud $X = \{x_i\}_{i=1}^N$ and semantic features $F = \{f_i\}_{i=1}^N$ extracted by a Transformer encoder, each point is represented as:
$u_i = [x_i, f_i] \in \mathbb{R}^{3 + d_f}.$
To achieve class-level reversible control, the discrete category label $y \in \{1, \dots, C\}$ is represented as a continuous embedding vector $e_y \in \mathbb{R}^{d_y}$.  
We define a trainable class embedding matrix $E = [e_1, \dots, e_C]$, where each $e_y$ corresponds to a learnable parameter for class $y$.  
To inject semantic information, we further introduce textual embeddings $e_y^{(\text{text})}$ from pretrained language models (e.g., CLIP).  
The two embeddings are fused via a MLP:
$e_y = \text{MLP}\big( [\, e_y^{(\text{train})},\, e_y^{(\text{text})} \,] \big).$
The conditional flow model $f_\theta(\cdot | e_y)$ learns a bijective mapping between input and latent spaces:
$z_i = f_\theta(u_i | e_y)$, $u_i = f_\theta^{-1}(z_i | e_y)$,
where $z_i \sim \mathcal{N}(0, I)$.  
This ensures both reversibility and information preservation.  
The Flow model consists of multiple affine coupling layers, where each layer performs a reversible transformation between two partitions of the input feature vector. In our framework, the Flow is feature-guided: each point representation $u_i = [x_i, f_i]$ is split into two parts, denoted by $(u_a, u_b)$, and transformed into $(v_a, v_b)$ as
$v_a = u_a,$
$v_b = u_b \odot \exp\!\big(s(u_a, e_y)\big) + t(u_a, e_y),
$
with the inverse transformation
$
u_a = v_a,
u_b = \big(v_b - t(v_a, e_y)\big)\odot \exp\!\big(-s(v_a, e_y)\big),
$
where $s(\cdot)$ and $t(\cdot)$ denote the scale and translation functions, respectively. 
In practice, both $s(\cdot)$ and $t(\cdot)$ are implemented as MLPs that take the first partition $x_1$ and the class embedding $e_y$ as inputs.  
Each MLP consists of hidden layers producing intermediate feature activations $h^{(l)}$:
$
h^{(l+1)} = \sigma \big(W^{(l)} h^{(l)} + b^{(l)} \big),
$
where $\sigma(\cdot)$ denotes a non-linear activation function GELU.
The coupling guarantees invertibility, and its log-likelihood is analytically computable. This design maintains strict reversibility while supporting end-to-end optimization.
\vspace{-3mm}
\begin{equation}
\mathcal{L}_{\text{NLL}} = -\log p(z) - \sum_i \log \left| \det \frac{\partial f_\theta(u_i|e_y)}{\partial u_i} \right|.
\vspace{-7mm}
\end{equation}



\begin{figure*}[t]
\begin{center}
\includegraphics[width=16cm, height=2.5cm]{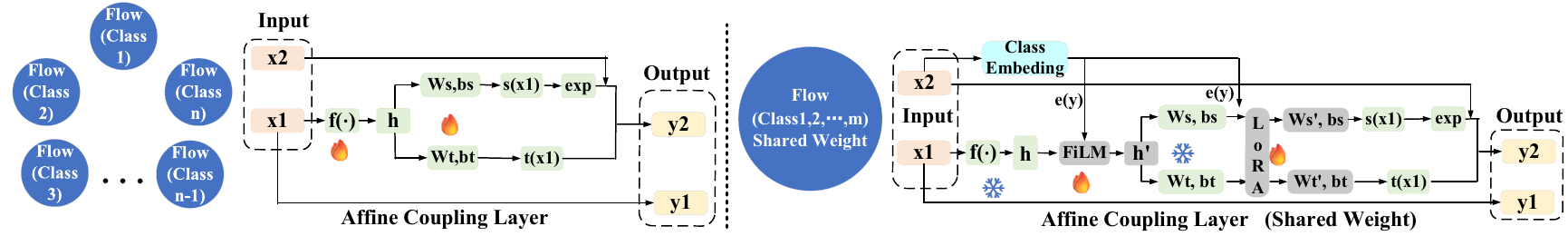}
\end{center}
\vspace{-4mm}
\caption{Comparison between the proposed parameter-efficient conditional Flow and the standard Glow model.
The proposed model (left) freezes the shared Flow backbone and introduces dual modulation for lightweight adaptation across classes: LoRA performs parameter-level low-rank fine-tuning, while FiLM provides feature-level semantic modulation. In contrast, the standard Glow (right) trains an independent Flow model for each class.
}
\vspace{-2mm}
\label{flowarch}
\end{figure*}

\subsection{Parameter Sharing}
In multi-class point cloud modeling, training an independent set of Normalizing Flow parameters for each category would result in a linear growth of model size, increasing both training and storage costs.  
Moreover, such an approach ignores the underlying geometric commonality shared across classes, making it difficult to achieve semantic-level generalization.
To address these issues, we propose a Dual Modulation Mechanism based on parameter sharing, built upon a shared Flow backbone; our architecture and the original Flow design are jointly illustrated in Fig.~\ref{flowarch}, enabling lightweight cross-class adaptation while maintaining strict reversibility of the Flow model. 
\textbf{(1) LoRA Parameter Modulation.}
In an invertible Flow, each affine coupling layer contains a scale network $s(\cdot)$ and a translation network $t(\cdot)$, parameterized by learnable weights $W_s$ and $W_t$. Training an independent pair $(W_s, W_t)$ for each class would cause the model size to grow linearly with the number of categories. To mitigate this issue, we adopt LoRA to enable parameter-efficient class adaptation by applying low-rank updates to the shared weights:
$
W_s' = W_s + \alpha_s(e_y)\, B_s A_s,
W_t' = W_t + \alpha_t(e_y)\, B_t A_t,
$
where $(A_s,B_s)$ and $(A_t,B_t)$ are globally shared low-rank matrices, and $\alpha_s(e_y)$ and $\alpha_t(e_y)$ are class-conditioned modulation coefficients predicted from the class embedding $e_y$ (lightweight MLP). This design achieves class-specific parameter adaptation through low-rank perturbations on shared weights, substantially reducing the overall parameter budget while retaining multi-class modeling capacity.
\textbf{(2) FiLM Feature Modulation.}
Beyond parameter adaptation, we further introduce FiLM to achieve semantic control at the feature level. Given an intermediate feature $h$ and the class embedding $e_y$, FiLM generates scaling and shifting coefficients $(\gamma,\beta)$ and applies feature-wise modulation:
$
h'=\gamma([h,e_y])\odot h+\beta([h,e_y]).
$
Within each affine coupling layer, FiLM is combined with LoRA to modulate the outputs of the scale and translation subnetworks. Let $u_a$ and $u_b$ denote the two partitions of the input point representation, and let $h_a$ denote the intermediate feature used for FiLM conditioning in the coupling subnetwork. 
\vspace{-1mm}
\begin{equation}
\small
\begin{aligned}
s'(u_a,e_y) &= \gamma_s([h_a,e_y]) \odot s_{\text{LoRA}}(u_a,e_y) + \beta_s([h_a,e_y]),\\
t'(u_a,e_y) &= \gamma_t([h_a,e_y]) \odot t_{\text{LoRA}}(u_a,e_y) + \beta_t([h_a,e_y]).
\end{aligned}
\vspace{-1mm}
\end{equation}

The resulting class-conditional affine coupling transformation is
\vspace{-1mm}
\begin{equation}
\small
\begin{aligned}
v_b = {} & u_b \odot \exp\!\big(s'(u_a,e_y)\big) + t'(u_a,e_y).
\end{aligned}
\vspace{-1mm}
\end{equation}
This design preserves the affine coupling form, and thus maintains strict invertibility while enabling class control at both parameter and feature levels.

\textbf{Dual-Level Scale Control:}
Since there exist significant geometric and semantic scale differences among classes and instances, direct reversible mapping may lead to latent space degradation and unstable decoding.  
We introduce a dual-level scale control module to ensure consistency and reversibility.  
For each point, the joint scale factor is defined as:
\vspace{-2mm}
\begin{equation}
s_i = \alpha_g(e_y)\|x_i - \mu_X\| + \alpha_s(e_y)\|f_i - \mu_F\|,
\vspace{-2mm}
\end{equation}
where $\alpha_g, \alpha_s$ are class-dependent geometric and semantic scaling weights, and $\mu_X, \mu_F$ are sample means.  
Inputs as:
$
\tilde{x}_i = \frac{x_i - \mu_X}{s_i}, \quad \tilde{f}_i = \frac{f_i - \mu_F}{s_i},
$
ensuring comparability across categories in a dimensionless space.  
During the inverse transformation:
$
x_i = s_i \cdot \tilde{x}_i + \mu_X, \quad f_i = s_i \cdot \tilde{f}_i + \mu_F,
$
thus restoring the  geometric structure.

\textbf{Semantic-Contrastive and Dispersion Regularization:}
To further enhance structural consistency of the class-conditional latent space, we introduce a semantic contrastive loss and a class embedding dispersion regularizer.  
The former enforces intra-class compactness and inter-class separation, while the latter prevents embedding collapse and ensures disentangled class modulation directions.
For each latent variable $z_i = f_\theta(u_i | e_{y_i})$, the class center as:
$
    c_y = \frac{1}{N_y} \sum_{i:y_i=y} z_i.
$

The semantic contrastive loss is:
\vspace{-1mm}
\begin{equation}
\small
\mathcal{L}_{\text{contrast}} = - \sum_i \log \frac{\exp(\text{sim}(z_i, c_{y_i}) / \tau)}{\sum_{y'} \exp(\text{sim}(z_i, c_{y'}) / \tau)}.
\vspace{-1mm}
\end{equation}
Class embedding dispersion regularization  as:
\vspace{-1mm}
\begin{equation}
\small
\mathcal{L}_{\text{disp}} = \frac{1}{C(C-1)} \sum_{i \neq j} \max(0, \tau_e - \|e_{y_i} - e_{y_j}\|_2^2).
\vspace{-1mm}
\end{equation}
These regularizers shape the CNF model to form reversible latent distributions that are compact within classes and separable across classes, achieving semantically consistent geometric mappings.

\subsection{Diffusion Flow Alignment}

Although our Flow model \( f_\theta(x \mid e_y) \) possesses strict invertibility and explicit density estimation capability, its latent space lacks explicit semantic constraints, leading the class-conditional distributions \(p_\theta(z \mid y)\) to exhibit unstable intra-class structures and overlapping inter-class regions, which in turn weakens both the separability of point cloud representations and the robustness of the encryption process.
To address this issue, we propose a {Diffusion–Guided Flow Alignment (DGFA)} framework,  
where the Diffusion model acts as a semantic regularizer for the Flow model.  
Through a temporally continuous noise diffusion process, the Diffusion module models the {dynamic manifold}  
of the latent distribution: it not only captures intra-class structural continuity  
but also implicitly learns the inter-class distribution gradient field (score field).  
Consequently, the Diffusion model learns a globally smooth and semantic distribution prior  
\(p_\phi(z|y)\), which serves as a {structural regularizer} for the Flow latent space.  
Specifically, we first train a Flow model to obtain the latent representation
$z = f_\theta(x|e_y),$
and then train a class-conditional Diffusion model \(p_\phi(z|y)\) within this Flow latent space  
to learn and stabilize the class-conditional structure of the latent distributions.  
This design anchors the Diffusion process within the Flow’s invertible mapping domain,  
thereby providing smoothness and semantic kernel alignment as regularization constraints.  
The forward and reverse diffusion processes are formulated as:
\vspace{-2mm}
\begin{equation}
z_0 = f_\theta(x|e_y), \quad 
z_t = \sqrt{\alpha_t} z_0 + \sqrt{1 - \alpha_t}\,\epsilon, 
\vspace{-2mm}
\end{equation}
where $\epsilon \sim \mathcal{N}(0, I)$,\(z_0\) denotes the latent variables produced by the Flow model,  
and \(z_t\) represents the noisy intermediate states.  
Thus, the Diffusion learning domain is strictly confined within the Flow’s latent space,  
ensuring consistency between the two models.  
The Diffusion model aims to predict the injected noise \(\epsilon_\phi(z_t, t, e_y)\),  
thereby learning the structural manifold of the Flow’s latent distribution:  
$
\mathcal{L}_{\text{diff}} =
\mathbb{E}_{z_0, t, \epsilon, y}
[\|\epsilon - \epsilon_\phi(z_t, t, e_y)\|_2^2].
$

\textbf{Training Strategy: }
To prevent the Diffusion module from interfering with the early convergence of the Flow model,  
we adopt a {three-stage progressive training strategy}  
to achieve stable co-optimization of geometric and semantic structures.  
In the {first stage},  
the Flow model \(f_\theta(x|e_y)\) is independently pretrained  
to establish a stable and reversible geometric mapping between point clouds and latent space.  
In the {second stage},  
the parameters of the Flow model are frozen,  
and a class-conditional Diffusion model \(p_\phi(z|y)\) is trained in the Flow’s latent space.  
Through the noise perturbation and denoising reconstruction processes,  
the Diffusion model learns a smooth and semantically consistent class-conditional distribution,  
forming a global structural prior for the latent space.  
In the {third stage},  
joint fine-tuning is performed by introducing a {Distribution Alignment Loss}  
to enforce probabilistic consistency between Flow and Diffusion.  
Specifically, we employ the symmetric KL divergence as the alignment objective:
\begin{equation}
\footnotesize
\mathcal{L}_{\text{align}} =
\left[
D_{\text{KL}}(p_\theta(z|y)\,\|\,p_\phi(z|y))
+ D_{\text{KL}}(p_\phi(z|y)\,\|\,p_\theta(z|y))
\right].
\end{equation}
where, \(p_\theta(z|y)\) denotes the class-conditional latent distribution modeled by Flow,  
while \(p_\phi(z|y)\) represents the semantic kernel distribution learned by Diffusion  
within the same latent space.  
This alignment mechanism forces the latent structure of the Flow model to gradually converge toward the smooth semantic kernels defined by the Diffusion model. Although Diffusion is trained on latent samples from Flow, its objective based on global noise smoothing and score alignment induces a distribution in latent space that is smoother and more semantically organized than Flow itself, and regularizes Flow through the alignment loss, thereby improving semantic consistency and linear separability while preserving geometric invertibility.

\subsection{Category Transition Energy}
Building on the latent space and parameter sharing established by the conditional Flow model—further aligned via diffusion—we introduce a measurable criterion to assess semantic/geometric proximity between classes and to guide source to target pairing and evolution. To this end, we define the Category Transition Energy to quantify the migration cost in latent space from class a to b: intuitively, the energy is lower when the two classes are more similar in latent distribution and geometry, and higher when they differ substantially or exhibit pronounced geometric distortion.

\textbf{Integral Form Definition. }
Let the latent energy field be defined as 
$
    E_\phi(z, y) = -\log p_\phi(z|y),
$
where $p_\phi(z|y)$ denotes the latent distribution of class $y$ learned by the diffusion model.  
Then, the energy transition from class $a$ to $b$ as a path integral over the energy gradient:
\vspace{-1mm}
\begin{equation}
\small
\label{Transition Energy}
\mathcal{E}_{a \rightarrow b}
= \mathbb{E}_{z \sim p_\phi(z|y_a)}
\!\left[
\int_0^1 
\|\nabla_z E_\phi(\gamma(t), y_t)\|_2^2 \, dt
\right],
\end{equation}
where $E_\phi(z, y) = -\log p_\phi(z|y)$ is the latent energy potential defined by the diffusion model, 
$\gamma(t)$ denotes the continuous latent trajectory from class $a$ to $b$, and $y_t$ is the interpolated class label along the path.
This integral measures the expected squared gradient of the energy potential along the trajectory, representing the ``energy flow cost'' between two latent distributions.
\textbf{Center–Distribution Approximation}
Since directly computing the path integral is computationally expensive, it can be approximated under two reasonable assumptions:
{Local Quadratic Assumption.}  
The energy function of class $y$ around its latent center can be approximated by a quadratic potential:
\vspace{-2mm}
\begin{equation}
E_\phi(z, y) \approx 
\tfrac{1}{2}(z - \mu_y)^\top \Sigma_y^{-1}(z - \mu_y),
\vspace{-2mm}
\end{equation}
where $\mu_y$ and $\Sigma_y$ denote the mean and covariance of class $y$ in the latent space, respectively.
\textbf{Linear Transition Path.} 
The trajectory between categories can be approximated as a linear interpolation:
$\gamma(t) = (1 - t)\mu_a + t\mu_b.
$
Under these conditions, the gradient term can be written as;
\vspace{-2mm}
\begin{equation}
\nabla_z E_\phi(\gamma(t), y_t) \approx \Sigma_y^{-1}(\gamma(t) - \mu_y).
\vspace{-2mm}
\end{equation}
Squaring and integrating along the path yields a term proportional to the geometric center difference:
$
\int_0^1 
\|\nabla_z E_\phi(\gamma(t), y_t)\|_2^2 dt 
\approx 
\|\mu_a - \mu_b\|_2^2.
$

\textbf{Covariance Stability and Distributional Difference.}  
If the covariance difference between classes is small, its variation can be absorbed as a distributional regularization term, measured by the Kullback–Leibler divergence:
\begin{equation}
\footnotesize
D_{\mathrm{KL}}\!\left(p_\phi(z|y_a)\,\|\,p_\phi(z|y_b)\right)
= 
\int 
p_\phi(z|y_a)
\log
\frac{p_\phi(z|y_a)}{p_\phi(z|y_b)} 
\, dz.
\end{equation}
This term captures the shape discrepancy between the latent distributions of different categories as learned by the diffusion model.
Combining the above approximations, the Category Transition Energy defined as:
\begin{equation}
\small
{
\mathcal{E}_{a \rightarrow b}
\approx
\|\mu_a - \mu_b\|_2^2
+ \lambda_E \,
D_{\mathrm{KL}}\!\left(p_\phi(z|y_a)\,\|\,p_\phi(z|y_b)\right)
}
\end{equation}
where $\lambda_E$ is a weighting coefficient that balances the relative influence of geometric center difference and distributional discrepancy.
$\|\mu_a - \mu_b\|_2^2$ represents the geometric distance between latent centers,
while the $D_{\mathrm{KL}}$ captures the distributional energy difference between category shapes.

\subsection{Category Transition Energy Approximation}

To better understand the definition of inter-class energy, this section derives a closed-form approximation of the category transition energy from a path integral formulation, expressed as the sum of a {center difference} and a {distribution difference} term.
We first define the complete integral form of the 
{Category Transition Energy} in Equ \ref{Transition Energy}.

\textbf{ Local Quadratic Expansion and Gradient Form: }
Under the local smoothness assumption in the latent space, the energy function can be approximated by a quadratic form:
\vspace{-2mm}
\begin{equation}
E_\phi(z, y) \approx \tfrac{1}{2} (z - \mu_y)^\top \Sigma_y^{-1} (z - \mu_y),
\vspace{-2mm}
\end{equation}
where $\mu_y$ and $\Sigma_y$ denote the mean and covariance of class $y$ in the latent space, respectively.
The gradient term is thus:
$
\nabla_z E_\phi(z, y) = \Sigma_y^{-1}(z - \mu_y).
$
Substituting into the integral yields:
\vspace{-3mm}
\begin{equation}
\|\nabla_z E_\phi(\gamma(t), y_t)\|_2^2 
= (\gamma(t) - \mu_{y_t})^\top \Sigma_{y_t}^{-2} (\gamma(t) - \mu_{y_t}).
\end{equation}

\textbf{Emergence of the Center Difference Term:}
If the covariance matrix $\Sigma_{y_t}$ varies only slightly along the path, 
it can be treated as a constant or its expectation $\bar{\Sigma}$.
Then, the path integral can be approximated as:
\vspace{-3mm}
\begin{equation}
\label{ua}
\int_0^1 \|\nabla_z E_\phi(\gamma(t), y_t)\|_2^2 \, dt
\approx (\mu_a - \mu_b)^\top \bar{\Sigma}^{-2} (\mu_a - \mu_b) .
\end{equation}
Equ \ref{ua} $\propto \|\mu_a - \mu_b\|_2^2.$
This yields the \textbf{center difference term}, which represents the geometric energy distance between class means in the latent space.
\textbf{Covariance Variation Term and Distribution Difference Approximation: }
To further account for the variation of the covariance $\Sigma_{y_t}$ along the trajectory, 
the energy integral can be expanded as:
\vspace{-2mm}
\begin{equation}
\int_0^1 \mathrm{tr}\!\left[ \Sigma_{y_t}^{-2} (\Sigma_a - \Sigma_b) \right] dt
+ (\mu_a - \mu_b)^\top \Sigma_{y_t}^{-1} (\mu_a - \mu_b).
\end{equation}
The first term reflects energy shifts induced by covariance variation, while the second term corresponds to the geometric energy difference caused by mean displacement.
However, directly computing the above integral is expensive and estimating the continuous covariance variation along the path is difficult.
Therefore, we introduce a statistically tractable and semantically equivalent measure—the \textbf{Kullback–Leibler (KL) divergence}—to quantify the overall discrepancy between the two latent distributions:
\begin{equation}
\small
D_{\mathrm{KL}}\!\left(p_\phi(z|y_a)\,\|\,p_\phi(z|y_b)\right) 
= \int p_\phi(z|y_a) \log \frac{p_\phi(z|y_a)}{p_\phi(z|y_b)} \, dz.
\end{equation}
Physically, this represents the expected ``extra energy cost'' incurred when samples from class $a$ are explained by the latent distribution of class $b$, 
i.e., it measures the {energy asymmetry} between the two distributions.
Assuming both latent distributions are approximately Gaussian, the KL divergence has a closed-form solution:
\vspace{-3mm}
\begin{equation}
\small
\begin{aligned}
D_{\mathrm{KL}}\!\left(p_\phi(z|y_a)\,\|\,p_\phi(z|y_b)\right) 
= \tfrac{1}{2}\! \bigg[ \,
\mathrm{tr}(\Sigma_b^{-1}\Sigma_a) \\
+ (\mu_b-\mu_a)^\top \Sigma_b^{-1}(\mu_b-\mu_a)
- k + \log \frac{\det\Sigma_b}{\det\Sigma_a} 
\bigg].
\end{aligned}
\end{equation}

\noindent where $k$ denotes the dimensionality of the latent space.
In this expression:
 $\mathrm{tr}(\Sigma_b^{-1}\Sigma_a)$ captures the covariance-induced energy difference;
  $(\mu_b - \mu_a)^\top \Sigma_b^{-1}(\mu_b - \mu_a)$ corresponds to the geometric energy from mean displacement;
 the remaining terms are dimensional and normalization corrections.
Thus, the covariance variation component can be absorbed into a KL-based distribution difference regularization term, 
replacing the original covariance energy offset in the integral, 
making the category transition energy more stable and statistically interpretable.

\textbf{Final Approximate Expression:}
Combining the above derivations, the \textbf{Category Transition Energy} can be approximated as:
\vspace{-2mm}
\begin{equation}
{
\mathcal{E}_{a \rightarrow b} 
\approx 
\|\mu_a - \mu_b\|_2^2 
+ \lambda_E \, D_{\mathrm{KL}}\big(p_\phi(z|y_a)\,\|\,p_\phi(z|y_b)\big)
}
\end{equation}
where $\lambda_E$ is a balancing coefficient controlling the relative contributions 
of geometric center difference and distributional shape discrepancy.
     The first term $\|\mu_a - \mu_b\|_2^2$ corresponds to the geometric center offset of the energy field;
   The second term $D_{\mathrm{KL}}$ captures the distributional energy difference induced by covariance variation.
This hybrid form serves as a computationally tractable approximation of the original energy integral,
preserving geometric interpretability while reflecting semantic distributional variation—
an effective formulation for measuring class-wise 
energy distance under the Flow–Diffusion framework.

\subsection{Category Energy Graph}
\textbf{Category Energy Graph: }
After computing the category transition energy $\mathcal{E}_{a \to b}$,  
all inter-class energy relations can be organized into a {directed weighted graph}: $\mathcal{G} = (\mathcal{V}, \mathcal{E}),$
where the node set $\mathcal{V} = \{v_1, v_2, \dots, v_C\}$ represents all categories,  
and the directed edge set $\mathcal{E} \subseteq \mathcal{V} \times \mathcal{V}$ indicates possible category transitions.  
Each directed edge $(v_i, v_j)$ is assigned a weight:
$ w_{ij} = \mathcal{E}_{i \to j},$
representing the energy required for transitioning from category $i$ to category $j$.
Due to the inherent asymmetry of the KL divergence in energy:
\begin{equation}
\small
D_{\mathrm{KL}}\!\left(p_\phi(z|y_i)\,\|\,p_\phi(z|y_j)\right)
\neq 
D_{\mathrm{KL}}\!\left(p_\phi(z|y_j)\,\|\,p_\phi(z|y_i)\right),
\end{equation}
it follows that
$\mathcal{E}_{i \to j} \neq \mathcal{E}_{j \to i}.$
Hence, the system is formulated as an {asymmetric energy graph}.
Each edge in the latent space represents a {one-way energy flow} between categories, reflecting the {irreversibility} of energy propagation.  
For computational convenience, all inter-category energy relationships are organized into an {asymmetric energy matrix} $\mathbf{E}$,  
where each row denotes the energy distribution from one category to all others,  
and each column represents the energy accumulation received from all other categories.

\textbf{Solving for the Optimal Mapping: }
Once the energy graph is constructed, the next goal is to find, for each category $i$,  
a unique target category $j^*$ that minimizes the transition energy without repetition across the entire set.  
Formally, this can be expressed as:
$
\min_{\pi \in \mathcal{S}_C} \sum_{i=1}^{C} \mathcal{E}_{i \to \pi(i)},$
where $\pi$ denotes a mapping permutation among categories, and $\mathcal{S}_C$ is the set of all possible permutations.  
This optimization corresponds to the {minimum-weight perfect matching problem}.
To efficiently obtain the globally optimal matching, the {Hungarian Algorithm} is employed,  
with a time complexity of $\mathcal{O}(C^3)$, ensuring an exact optimal solution.  
The algorithm takes the energy matrix $\mathbf{E}$ as input and outputs an optimal {matching matrix} $\mathbf{M}$ defined as a binary mapping matrix $\mathbf{M} \in \{0,1\}^{C \times C}$, where $\mathbf{M}_{ij} = 1$ if category $i$ is mapped to category $j$, and $\mathbf{M}_{ij} = 0$ otherwise.
The matching matrix satisfies the constraints:
$\sum_{i}\mathbf{M}_{ij}$=1,
$\sum_{j}\mathbf{M}_{ij}$=1,
ensuring that each category is uniquely mapped to one target category without duplication.
Through the optimal matching result,  
the final energy mapping relationship  can be expressed as:
i $\rightarrow$ $\pi(i)$,
$\text{energy} = \mathcal{E}_{i \to \pi(i)}.$




\subsection{Category Transition Flow}
We established the globally optimal category matching set 
$\mathcal{M} = \{(y_a, y_b)\}$ through the construction of the category transition energy graph.  
We now aim to build a continuous, reversible, and structurally stable transformation mechanism in the latent space.  
To this end, we introduce a continuous flow modeling approach based on Ordinary Differential Equations (ODEs), which dynamically describes the category transition process while keeping the energy variation along the flow path as small as possible.
Traditional category transformations are often based on discrete mappings or conditional generation, 
which ignore the continuous geometric structure of the latent space.  
To achieve smooth inter-category transitions, we adopt an ODE-based continuous flow:
$
\frac{dz_t}{dt} = g_\psi(z_t, e_{y_t}), \quad t \in [0, 1],
$
where $z_t \in \mathbb{R}^d$ is latent variable at time $t$;
    $g_\psi(\cdot)$ is neural network–parameterized flow field function;
    $e_{y_t}$: the class-conditioned embedding, obtained via linear interpolation between the source and target embeddings:
   $
    e_{y_t} = (1 - t)e_{y_a} + t e_{y_b};
 $
    $t$ denotes continuous time parameter controlling the transition degree between categories.
By integrating over time, the continuous mapping from  $A$ to $B$:
\vspace{-3mm}
\begin{equation}
T_{a \to b}(z_a) = z_a + \int_0^1 g_\psi(z(t), e_{y_t})\,dt. 
\vspace{-3mm}
\end{equation}
This ODE formulation ensures {continuity}, {reversibility}, and {physical interpretability}, 
since the transformation can be exactly recovered via backward integration:
$T_{b \to a}(T_{a \to b}(z_a)) = z_a. $
However, without additional physical constraints, 
the unconstrained neural flow field $g_\theta$ may cause undesired energy drift or distribution shift, 
as different categories may have distinct energy landscapes, leading to semantic drift or structural distortion.  
Therefore, we introduce a \textbf{local energy smoothness constraint} to regularize the flow dynamics.
Given $E_\phi(z, y)$, we require that during category transitions,
the energy variation along the trajectory remains minimal:
\vspace{-3mm}
\begin{equation}
\frac{dz}{dt} = g_\psi(z, e_{y_t}), 
\quad \text{s.t. } E_\phi(z(t), y_t) \approx \text{const.} 
\vspace{-3mm}
\end{equation}
This condition implies that, during the transformation from category $A$ to $B$, 
the sample’s energy in latent space should change as little as possible. In other words, the direction of $g_\psi$ should be approximately orthogonal to the energy gradient field $\nabla_z E_\phi$, 
allowing the flow to slide smoothly along iso-energy surfaces.  
This ensures that the category transition adheres to the principle of {minimal energy perturbation}.
\vspace{-2mm}
\begin{equation}
\small
\frac{d}{dt}E_\phi(z_t, y_t)
= \nabla_z E_\phi(z_t, y_t)^\top g_\psi(z_t, e_{y_t}) \approx 0. 
\vspace{-1mm}
\end{equation}

\textbf{Continuous Category Mapping:}
Under the above constraint, the continuous category transformation is defined:
\vspace{-1mm}
\begin{equation}
\small
\label{weifenzhuanhua}
\begin{aligned}
T_{a \to b}(z_a)
&= z_a + \int_0^1 g_\psi(z(t), e_{y_t})\,dt, \\
&\quad \text{s.t. } 
\nabla_z E_\phi(z_t, y_t)^\top g_\psi(z_t, e_{y_t}) \approx 0.
\end{aligned}
\vspace{-2mm}
\end{equation}
During inference, the original category can be recovered by backward integration:
\vspace{-3mm}
\begin{equation}
\label{dechangweifen}
T_{b \to a}(z_b)
= z_b - \int_0^1 g_\psi(z(t), e_{y_{1-t}})\,dt, 
\vspace{-3mm}
\end{equation}
thus satisfying strict reversibility:
$T_{b \to a}(T_{a \to b}(z_a))$ = $z_a. $
That ensures that the category transition follows continuous dynamic laws 
while maintaining energy balance throughout the process.

\subsubsection{Energy Conservation Constraint }

In the reversible flow modeling of category transitions, the evolution of the latent space is driven by the neural flow field $g_\theta(z_t, e_{y_t})$.  
However, due to structural discrepancies among category distributions, different categories often correspond to distinct energy landscapes $E_\phi(z, y)$.  
Without proper constraints, the model may produce local energy oscillations, trajectory distortions, or gradient explosions during inter-category mapping, 
which can break reversibility and global stability.  
To address this, we introduce an \textbf{energy conservation constraint} during training to regulate the relationship between the flow direction and the energy gradient, 
ensuring that the latent dynamics remain balanced and smooth in the energy space.

\textbf{Motivation.}
During the transition from category $y_a \to y_b$, the category energy function $E_\phi(z, y)$ characterizes the statistical potential landscape of samples in the latent space.  
Since the energy difference between categories inherently exists, i.e.,
$
E_\phi(z, y_a) \neq E_\phi(z, y_b),
$
the category transformation must involve global energy variation.  
However, the purpose of the energy conservation constraint is {not} to eliminate this difference, but rather to ensure that the transition path itself does not introduce extra energy dissipation or artificial energy amplification.  
In other words, the model should perform smooth {energy-level sliding} along iso-energy surfaces, instead of chaotic traversal across different energy layers.
Let $E_\phi(z_t, y_t)$ denote the potential energy of latent state $z_t$ under category $y_t$.  
Its time derivative is:
\vspace{-2mm}
\begin{equation}
\begin{aligned}
\frac{d}{dt} E_\phi(z_t, y_t)
&= \nabla_z E_\phi(z_t, y_t)^\top \frac{dz_t}{dt} \\
&= \nabla_z E_\phi(z_t, y_t)^\top g_\theta(z_t, e_{y_t}).
\end{aligned}
\vspace{-3mm}
\end{equation}

To achieve energy conservation, we constrain this temporal derivative to approach zero throughout the flow process:
\vspace{-2mm}
\begin{equation}
\nabla_z E_\phi(z_t, y_t)^\top g_\theta(z_t, e_{y_t}) \approx 0. 
\vspace{-2mm}
\end{equation}

Accordingly, the \textbf{energy conservation regularization term} is defined as:
\vspace{-2mm}
\begin{equation}
\mathcal{L}_{\text{energy}} 
= \mathbb{E}_{z_t} \big[ 
\big| \nabla_z E_\phi(z_t, y_t)^\top g_\theta(z_t, e_{y_t}) \big|
\big]. 
\vspace{-2mm}
\end{equation}
This regularization suppresses the component of the flow field along the energy gradient direction, 
forcing samples to move primarily along iso-energy surfaces, thereby maintaining local equilibrium in the latent energy structure.
From a latent-space geometric perspective, different category distributions $p(z|y)$ can be viewed as manifolds lying on distinct energy surfaces $E_\phi(z,y) = c_y$.  
If the direction of $g_\theta$ is not orthogonal to the energy gradient $\nabla_z E_\phi$, the flow trajectories will frequently cross energy layers, 
causing disordered fluctuations and density compression in the energy domain.  
The model is encouraged to learn energy-level sliding directions that satisfy:
$
g_\theta(z_t, e_{y_t}) \perp \nabla_z E_\phi(z_t, y_t), 
$
allowing samples to deform continuously along {work-free paths} on the energy landscape.  
This flow behavior preserves the stability of the energy structure and prevents geometric distortion or density collapse in the latent evolution.

\textbf{Effect on ODE Integration.}
The energy conservation constraint acts as a smooth regularizer in the ODE integration process.  
When $g_\theta$ frequently aligns with $\nabla_z E_\phi$, the local gradient magnitude $\|\nabla_z E_\phi\|$ becomes excessively large, 
causing abrupt step-size variations in the ODE solver.  
This leads to unstable adaptive step-size reduction, reduced training efficiency, and compromised reversibility.  
By applying the constraint, the interaction between $\|g_\theta\|$ and $\nabla_z E_\phi$ is smoothed, 
thereby improving the following properties:
\textbf{Integration Stability:} Reduced energy fluctuations allow consistent step sizes in both forward and backward integration.
\textbf{Gradient Smoothness:} Prevents the flow field from exhibiting gradient explosion or vanishing between energy layers.
  \textbf{Reversible Consistency:} Ensures that backward integration trajectories numerically reconstruct the forward paths, 
    enhancing the overall reversibility of the mapping.

\subsubsection{Neural Flow Field}
The flow field function $g_\psi$ is implemented via an MLP
that learns the free-flow dynamics in the latent space.  
Its input consists of the current latent variable $z_t$, category embedding $e_{y_t}$, and time $t$:
\vspace{-3mm}
\begin{equation}
g_\psi : (z_t, e_{y_t}, t) \mapsto v_t = \frac{dz_t}{dt}. 
\vspace{-3mm}
\end{equation}
This neural structure learns a vector field satisfying the energy conservation constraint,
ensuring that sample trajectories slide along inter-category energy manifolds rather than jumping between them.
{Distribution Alignment Loss.}
\vspace{-2mm}
\begin{equation}
\small
\mathcal{L}_{\text{dist}} 
= \left\|
\mathbb{E}_{z_a \sim p_{z_a}}[\phi(T_{a\to b}(z_a))]
- \mathbb{E}_{z_b \sim p_{z_b}}[\phi(z_b)]
\right\|_2^2, 
\end{equation}
where $\phi(\cdot)$ denotes the embedding function that measures the discrepancy 
between the transformed and target distributions.
{Reversibility Consistency Loss.}
This term ensures strict numerical reversibility of the bidirectional mappings:
\vspace{-2mm}
\begin{equation}
\mathcal{L}_{\text{rev}} 
= \mathbb{E}_{z_a\sim p_{z_a}}
\left[
\|T_{b\to a}(T_{a\to b}(z_a)) - z_a\|_2^2
\right]. 
\end{equation}
Local Energy Smoothness Loss is as:
\vspace{-2mm}
\begin{equation}
\mathcal{L}_{\text{energy}} 
= \mathbb{E}_{z_t}
\big[
|\nabla_z E_\phi(z_t, y_t)^\top g_\psi(z_t, e_{y_t})|
\big], 
\vspace{-2mm}
\end{equation}
which constrains the neural flow field to be nearly orthogonal to the energy gradient, 
preventing oscillations or collapses caused by excessive energy dependence.  
This ensures minimal energy variation, yielding a smooth, reversible, and physically consistent flow.
The weighting coefficients $\lambda_1, \lambda_2, \lambda_3$ 
control the relative importance of distribution alignment, reversibility, and energy balance.  
The locally energy-conservative category transition flow maintains continuous and minimally perturbed energy evolution in the latent space:
$\frac{d}{dt}E_\phi(z_t, y_t) \approx 0, 
E_\phi(z_1, y_b) - E_\phi(z_0, y_a)$  is smooth and controlled. 
This formulation realizes a {continuous transformation along energy surfaces}, 
making the category transition semantically coherent, geometrically reversible, 
and stable.

\subsubsection{ Neural Flow Field Fitting}

During the category transition stage, the latent distributions of categories $y_a$ and $y_b$ have already been globally aligned through category energy matching. 
To further enable reversible mapping at the sample level, we introduce a neural network–parameterized flow field function $g_\theta$, 
which learns the temporal evolution of latent variables in continuous time. 
This process can be viewed as fitting a time-continuous, differentiable, and invertible dynamical system in the latent space.

We define the latent flow equation as:
\vspace{-3mm}
\begin{equation}
\frac{dz_t}{dt} = g_\theta(z_t, e_{y_t}), \quad t \in [0, 1],
\vspace{-3mm}
\end{equation}
where:
 $z_t \in \mathbb{R}^d$: the latent state at time $t$;
     $g_\theta(\cdot)$: the neural network–parameterized velocity field function;
     $\theta$: learnable network parameters;
     $\tfrac{dz_t}{dt}$: the instantaneous flow velocity in the latent space.
The objective of the neural network is to learn a continuous and reversible vector field such that, when integrated from $t=0$ (category $A$) to $t=1$ (category $B$), the pushed-forward distribution satisfies:
$
p_{z_b} = (T_{a\to b})_{} p_{z_a}, 
z_b = z_a + \int_0^1 g_\theta(z_t, e_{y_t})\, dt.$

\textbf{Network Architecture.}
The neural flow field $g_\theta$ is implemented as a MLP to ensure continuity and differentiability of the input–output mapping.  
The input consists of the current latent variable $z_t$, the conditional embedding $e_{y_t}$, and the scalar time $t$, forming:
$
g_\theta: (z_t, e_{y_t}, t) \mapsto v_t,
$
where $v_t = \tfrac{dz_t}{dt}$ denotes the local flow direction and speed of the sample in the latent space.
The network includes three linear mappings and several residual modules.
The first layer performs a linear transformation and layer normalization on the input vector $[z_t, e_{y_t}, t] \in \mathbb{R}^{d + d_y + 1}$;
  The second layer continues hidden-to-hidden mapping to capture nonlinear potential structures in the latent field;
$N$ residual blocks are stacked to stabilize gradient propagation and maintain the smoothness of the continuous flow field;
The final output layer projects features into $\mathbb{R}^d$, producing the flow vector $g_\theta(z_t, e_{y_t})$, 
    which describes local dynamic behavior in the latent space.
\textbf{Training Objectives.}
The parameters of $g_\theta$ are jointly optimized by minimizing the following combined objective:
\vspace{-2mm}
\begin{equation}
\mathcal{L}_{\text{total}} = \lambda_1 \mathcal{L}_{\text{dist}} + \lambda_2 \mathcal{L}_{\text{rev}} + \lambda_3 \mathcal{L}_{\text{energy}},
\vspace{-2mm}
\end{equation}

\textbf{Distribution Alignment Loss.}
\vspace{-2mm}
\begin{equation}
\begin{aligned}
\mathcal{L}_{\text{dist}} 
&= \mathrm{MMD}\!\left( T_{a\to b}(z_a),\, z_b \right) \\[4pt]
&= \left\| 
\mathbb{E}_{z_a}[\phi(T_{a\to b}(z_a))] 
- \mathbb{E}_{z_b}[\phi(z_b)]
\right\|_2^2.
\end{aligned}
\vspace{-2mm}
\end{equation}
which enforces transformed distribution to align with the target category’s latent distribution.

\textbf{ Reversibility Constraint.}
\vspace{-2mm}
\begin{equation}
\mathcal{L}_{\text{rev}} 
= \mathbb{E}_{z_a}\|T_{b\to a}(T_{a\to b}(z_a)) - z_a\|_2^2,
\vspace{-2mm}
\end{equation}
ensuring that $g_\theta$ remains bijective under bidirectional integration.

\textbf{Energy Conservation Constraint.}
\vspace{-2mm}
\begin{equation}
\mathcal{L}_{\text{energy}} 
= \mathbb{E}_{z_t}\!
\left[
\big| \nabla_z E_\phi(z_t, y_t)^\top g_\theta(z_t, e_{y_t}) \big|
\right],
\vspace{-2mm}
\end{equation}
which enforces orthogonality between the flow direction and the energy gradient, preventing local oscillations and trajectory drift.

\textbf{Optimization Process.}
During training, the neural flow field $g_\theta$ is fitted via the following steps:
\textbf{Sampling.}  
Sample starting and target points from the latent distributions of categories $A$ and $B$:
$
z_a \sim p_{z_a}, \quad z_b \sim p_{z_b}.
$
\textbf{Forward Integration.}  
Using an ODE solver (e.g., Dormand–Prince or Runge–Kutta), integrate based on the current network-predicted $g_\theta$:
$   
z_{t+\Delta t} = z_t + g_\theta(z_t, e_{y_t})\, \Delta t.
$
\textbf{Loss Computation.}  
Compute the transformed result $z_b' = T_{a\to b}(z_a)$, and evaluate 
$\mathcal{L}_{\text{dist}}$, 
$\mathcal{L}_{\text{rev}}$, 
and $\mathcal{L}_{\text{energy}}$.
\textbf{Backpropagation.}  
Compute gradients using the adjoint sensitivity method:
\vspace{-2mm}
\begin{equation}
\frac{d\mathcal{L}}{d\theta} = - \int_0^1 a_t^\top \frac{\partial g_\theta(z_t, e_{y_t})}{\partial \theta}\, dt,
\vspace{-2mm}
\end{equation}
where $a_t$ is the adjoint state defined as:
\vspace{-2mm}
\begin{equation}
\frac{da_t}{dt} = -a_t^\top \frac{\partial g_\theta(z_t, e_{y_t})}{\partial z_t}.
\vspace{-2mm}
\end{equation}
The parameters $\theta$ are updated using the Adam.
\textbf{Iteration Until Convergence.}  
When the distribution alignment loss converges and the bidirectional reconstruction error stabilizes, 
$g_\theta$ is considered successfully fitted to the latent flow from category $A$ to category $B$.
During optimization: $\mathcal{L}_{\text{dist}}$ drives alignment at the distributional level;  $\mathcal{L}_{\text{rev}}$ maintains numerical stability and reversibility;  $\mathcal{L}_{\text{energy}}$ enforces smoothness and local energy conservation.
Through their joint optimization, the neural network forms a smooth, energy-balanced, and reversible latent flow,
achieving a stable mapping between categories. 
After convergence, $g_\theta$ accurately captures the latent-space dynamics from category $A$ to category $B$.

\subsection{Encryption Decryption  Process}
\textbf{Encryption Process. }
In the encryption phase, the original point cloud $x_a$ and its extracted feature $f_a$ are combined into a unified representation $u_a = [x_a, f_a]$ , 
which is encoded by the reversible flow model into the latent space as:
$z_a = f_\theta(u_a | y_a),$
where $y_a$ denotes the original class label.
Subsequently, based on the globally optimal category mapping table $\mathcal{M}^*$ 
obtained through the energy graph, 
the corresponding target class is determined as:
$y_b = \mathcal{M}^*(y_a).$
To achieve latent-space migration from class $a$ to class $b$, 
the sample evolves along an \textit{energy-conservative flow} defined by Equ \ref{weifenzhuanhua}.
Finally, the transformed latent variable is decoded back into the point cloud space 
through the inverse mapping function $f_\theta^{-1}$:
$u_b = f_\theta^{-1}(z_b | y_b),$
yielding the encrypted point cloud $x_b$.

\textbf{Key Set Definition.}
Before performing decryption, the receiver must obtain the complete key set, which is defined as:
$\mathcal{K} = \{\theta, \psi,E_\theta,  \mathcal{M}^*\},$
where:
    $\theta$: parameters of the {normalizing flow}, responsible for the bijective mapping between the point cloud and the latent space;
   $\psi$: parameters of the {neural ODE flow}, governing the continuous dynamic evolution of different categories in the latent space;
   $E_\phi$: parameters of the {energy function};
   $\mathcal{M}^*$: the {category mapping table}, which determines the one-to-one correspondence between encryption and decryption categories.
   
\textbf{Decryption Process.}
The decryption process strictly follows the system's reversibility assumption, 
recovering the original point cloud data through time-reversed integration and inverse category mapping.
For the received encrypted point cloud $x_b$ and its feature $f_b$.
The $u_b = [x_b, f_b]$ is encoded by the same reversible flow model to obtain its latent variable:
$z_b = f_\theta(u_b | y_b),$
where the encoding process is identical to that used in the encryption phase 
and relies on the same flow parameters $\theta$.
Based on the known optimal category mapping table $\mathcal{M}^*$, 
the original category corresponding to the encrypted one is determined as:
$y_a = \mathcal{M}^{*-1}(y_b),$
ensuring semantic determinism of the decryption process.
By leveraging the reversibility of the ordinary differential equation (ODE), 
the latent variable is reconstructed via time-reversed integration Equ \ref{dechangweifen}.
Since the flow evolves under local energy-conservation constraints, 
the integration path remains stable and continuous, ensuring:
$T_{b\to a}(T_{a\to b}(z_a)) = z_a,$
which guarantees strict invertibility of the decryption process in the latent space.
The recovered latent variable $z_a$ is then decoded back into the point cloud space 
through the inverse mapping function of the reversible flow:
$u_a = f_\theta^{-1}(z_a | y_a),$
yielding the reconstructed original point cloud $x_a$.


\section{Experiment}
\subsection{Settings}

\textbf{ModelNet40.}
\cite{wu20153d} is a widely used benchmark for 3D object classification, consisting of 12,311 synthetic CAD models from 40 object categories.
With the official split providing 9,843 models for training and 2,468 for testing. 
\textbf{ScanObjectNN.}
\cite{uy2019revisiting} is a real-world indoor object dataset constructed from 3D scans of cluttered indoor scenes, such as those from SceneNN and ScanNet.
In contrast to synthetic CAD datasets, point clouds in ScanObjectNN are cropped from noisy scans and often exhibit background clutter, partial observations, and occlusions.
The benchmark contains approximately 15,000 objects from 15 categories, corresponding to 2,902 unique object instances, making it substantially more challenging than ModelNet40.
\textbf{ShapeNetPart.}
\cite{chang2015shapenet} is a part-level segmentation benchmark derived from the ShapeNetCore collection of 3D CAD models.
It contains 16,881 shapes from 16 object categories, where each point cloud is annotated with 2--6 semantic part labels, yielding more than 50 fine-grained part types in total.
Each shape typically consists of approximately 1,000 to 3,000 points.

\textbf{Transformed ShapeNetPart.}
Since ShapeNetPart adopts category-specific part annotations, semantically transformed point clouds do not have directly corresponding ground-truth part labels for the target category. Therefore, we select, for each category, the top four performing models from five segmentation architectures trained on the dataset, namely PointMLP, PointCNN, MaskFeat3D, PointNet++, and PointMeta, and use them as independent evaluators. For each transformed point cloud sample, we obtain four point-wise segmentation predictions. If at least three models assign the same segmentation label to a point, we regard that label as a high-consensus pseudo-label for that point. Otherwise, the point is treated as an inconsistent point, which is considered an unreliable region in the target semantic space and counted as a failed point in the segmentation statistics.
The category-specific ratios of inconsistent points are: aero 6\%, bag 10\%, cap 4\%, car 13\%, chair 2\%, earphone 15\%, guitar 5\%, knife 16\%, lamp 7\%, laptop 2\%, motor-bike 16\%, mug 4\%, pistol 12\%, rocket 25\%, skateboard 12\%, and table 10\%.

\begin{table*}[t]
\centering
\resizebox{1\textwidth}{!}{
\scriptsize
\setlength{\tabcolsep}{3pt}
\begin{tabular}{llccccccc}
\toprule
\textbf{Dataset} & \textbf{Method} & \textbf{PointNet \cite{qi2017pointnet}} & \textbf{PointNet++ \cite{qi2017pointnet++}} & \textbf{DGCNN \cite{wang2019dynamic}} & \textbf{PointCNN \cite{li2018pointcnn}} & \textbf{GBNet \cite{qiu2022geometric}} & \textbf{Point Trans \cite{zhao2021point}} & \textbf{PointMLP \cite{marethinking}} \\
\midrule
\textbf{ScanObjectNN} & Acc  & 0.682 / 0.664 & 0.779 / 0.759 & 0.781 / 0.771 & 0.785 / 0.775 & 0.805 /0. 785 & 0.828 / 0.808 & 0.854 / 0.840 \\
 & NCRR & 0.651 & 0.746 & 0.748 & 0.752 & 0.772 & 0.794 & 0.820 \\
 & OCLR & 0.172 & 0.120 & 0.119 & 0.117 & 0.106 & 0.094 & 0.080 \\
 & OLRR & 0.198 & 0.146 & 0.141 & 0.138 & 0.126 & 0.112 & 0.098\\
\midrule
\textbf{ModelNet40} & Acc  & 0.892 / 0.881 & 0.909 / 0.903 & 0.929 / 0.918 & 0.925 / 0.914 & 0.938 / 0.927 & 0.929 / 0.918 & 0.941 / 0.929 \\
 & NCRR & 0.850 & 0.860 & 0.875 & 0.872 & 0.885 & 0.876 & 0.890 \\
 & OCLR & 0.090 & 0.085 & 0.075 & 0.077 & 0.070 & 0.074 & 0.065 \\
 & OLRR & 0.112 & 0.105 & 0.096 & 0.092 & 0.085 & 0.082 & 0.078 \\
\bottomrule
\end{tabular}
}
\caption{Comparison of classification performance on the {ScanObjectNN \cite{uy2019revisiting}} and {ModelNet40} datasets \cite{wu20153d} .
In the acc column, the value before the slash represents the {accuracy on the original data},
while the value after the slash indicates the {acc obtained by training the model on the encrypted (category-transformed) data}.}
\vspace{-6mm}
\label{tab:scanobject-modelnet}
\end{table*}

\textbf{Evaluation Metrics.}
In the evaluation setting of this paper, we consider two types of attackers. The first is a fixed attacker, who can only use a classifier trained on the original data and directly apply it to the transformed point clouds to determine whether these samples still retain sufficiently strong cues of their original classes. The second is a stronger adaptive attacker, who not only has access to the transformed samples but can also retrain a new classifier on the transformed training data with the goal of recovering the original labels. Based on these two attack scenarios, we evaluate the proposed method from three aspects: target-class recognizability, original semantic leakage under fixed attacks, and original semantic recovery under adaptive attacks.We adopt the following metrics: New-Class Recognition Rate (NCRR), Original-Class Leakage Rate (OCLR), and Original Label Recovery Rate (OLRR).
We first train a standard classifier
on the original point cloud dataset and keep this classifier fixed in all experiments.
We then apply the class-transfer procedure to the test samples and compute both metrics
only on the transformed point clouds. Let the original class of the \(i\)-th sample be
\(y_i^{\text{orig}}\), and its corresponding target class in the class-transfer table be
\(y_i^{\text{tgt}}\). Denote by \(\hat{y}_i^{(\text{after})}\) the prediction of the fixed
classifier on the transformed sample. The New-Class Recognition Rate is defined as
\vspace{-3mm}
\begin{equation}
    \mathrm{NCRR}
= \frac{1}{N}\sum_{i=1}^{N}
\mathbf{1}\big[\hat{y}_i^{(\text{after})} = y_i^{\text{tgt}}\big],
\vspace{-3mm}
\end{equation}
which measures, from an attacker’s perspective, the proportion of transformed samples that
are classified as their designated target class, reflecting the utility of the class-transfer
procedure in effectively aggregating source samples into the target class.
The Original-Class Leakage Rate as 
\vspace{-3mm}
\begin{equation}
    \mathrm{OCLR}
= \frac{1}{N}\sum_{i=1}^{N}
\mathbf{1}\big[\hat{y}_i^{(\text{after})} = y_i^{\text{orig}}\big],
\vspace{-3mm}
\end{equation}
which quantifies the proportion of transformed samples that are still recognized as their
original class by the fixed classifier. This can be interpreted as a measure of residual
identity/semantic information leakage, where lower OCLR values indicate stronger privacy
protection.
To further evaluate whether the original class information in semantically encrypted point clouds can still be recovered by an adaptive attacker, we introduce the OLRR. We assume that the attacker has access to the encrypted training data and retrains a classifier using their corresponding original class labels as supervision. We then measure the success rate of recovering the original labels on the encrypted test set. Let \(x_i^{\mathrm{enc}}\) denote the \(i\)-th encrypted sample, \(y_i^{\mathrm{orig}}\) denote its corresponding original class label, and \(f_{\mathrm{adv}}\) denote the attack classifier retrained on the encrypted training set. The OLRR as
\vspace{-3mm}
\begin{equation}
\mathrm{OLRR} = \frac{1}{N}\sum_{i=1}^{N}\!\left[f_{\mathrm{adv}}\!\left(x_i^{\mathrm{enc}}\right)=y_i^{\mathrm{orig}}\right],
\vspace{-3mm}
\end{equation}
A lower OLRR indicates that less original semantic information remains in the encrypted samples.

\textbf{Inference:} Our method has low inference overhead. The diffusion module is used only during training for distribution alignment in the shared latent space and is not executed at inference time. In addition, the diffusion process operates in a low-dimensional latent space, which keeps its training cost within a manageable range. The construction of the class energy maps and the Hungarian matching are both offline procedures and only need to be precomputed once, introducing no online inference overhead.
From the perspective of scalability, the Flow module introduces only class-conditional modulation parameters, while latent diffusion models the shared latent space, so the additional overhead does not grow linearly with the number of classes. The Neural ODE at inference time operates only on category transfer in the latent space, and its computational complexity is independent of the total number of classes.
On an RTX 4090, the total inference time for encrypting and decrypting a single sample is about 100 ms, among which the ODE-based latent migration accounts for about 30 ms. During training, each epoch takes about 2 hours, and the model typically converges after around 20 epochs.
\begin{figure}[t]
\begin{center}
\includegraphics[width=7cm, height=6cm]{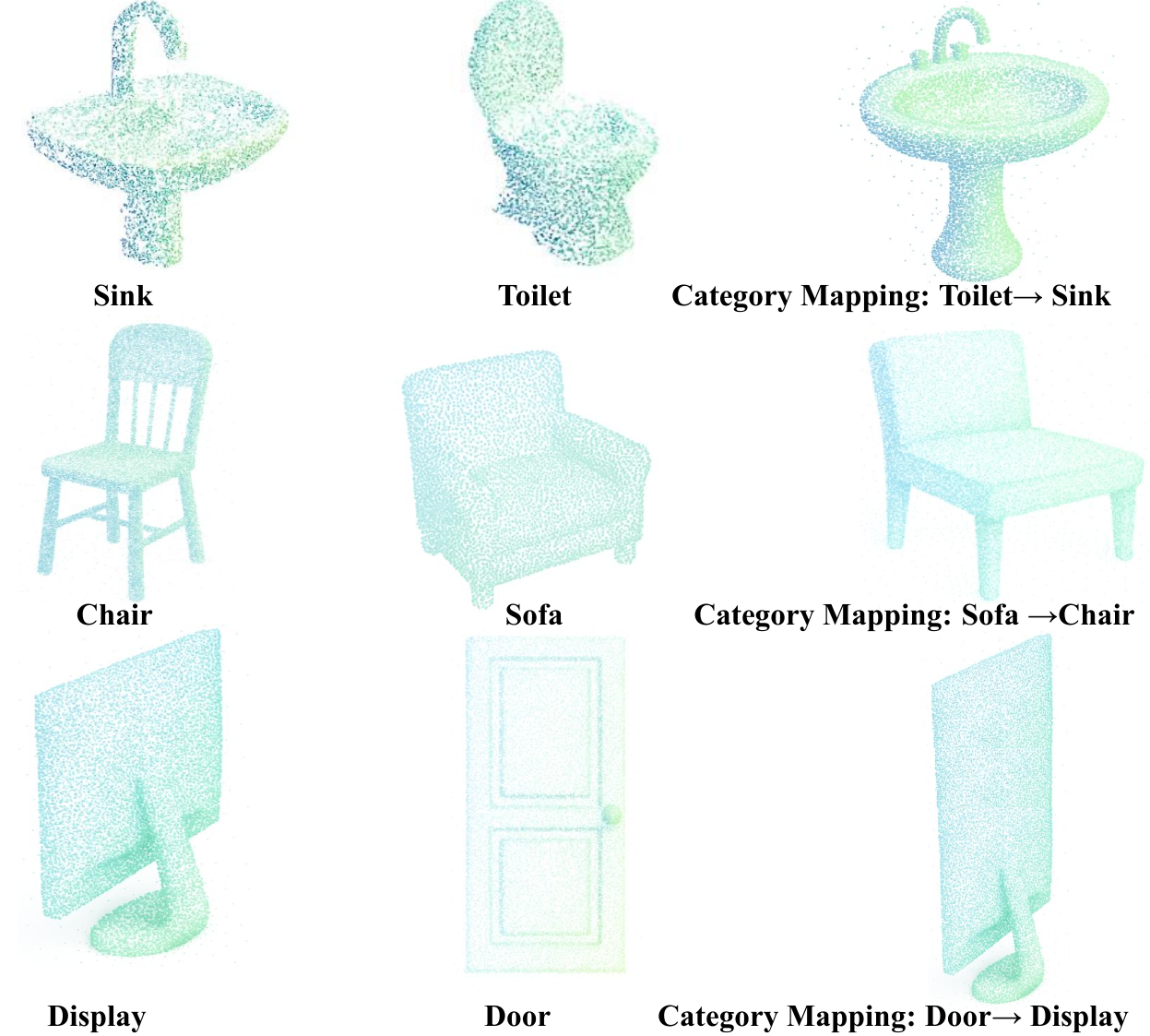}
\end{center}
\vspace{-4mm}
 \caption{Visual point clouds of class transfer on the {ScanObjectNN \cite{uy2019revisiting}}.  Each triplet shows, from left to right: the target, the source, and the result after encryption }
 \vspace{-2mm}
\label{transfer}
\end{figure}

\begin{table*}[t]
\centering
\resizebox{1\textwidth}{!}{
\scriptsize
\begin{tabular}{l|ccccccccccccccccc}
\toprule
Method & aero & bag & cap & car & chair & aerphone & guitar & knife & lamp & laptop & motor-bike & mug & pistol & rocket & skateboard & table & Mean \\
\midrule
PointNet & 83.4/81.53 & 78.7/80.0 & 82.5/84.27 & 74.9/73.18 & 89.6/87.83 & 73.0/71.19 & 91.5/89.97 & 85.9/84.06 & 80.8/82.51 & 95.3/93.99 & 65.2/63.31 & 93.0/94.47 & 81.2/79.34 & 57.9/56.48 & 72.8/71.02 & 80.6/78.71 & 81.2/79.95 \\
PointNet++ & 82.3/80.93 & 79.0/77.57 & 87.7/86.46 & 77.3/75.52 & 90.8/92.05 & 71.8/70.79 & 91.0/89.88 & 85.0/83.03 & 83.7/84.86 & 95.3/94.05 & 71.6/70.38 & 94.1/92.9 & 81.3/80.0 & 58.7/56.75 & 76.4/75.0 & 82.6/84.38 & 82.7/81.47 \\
Kd-Net & 74.3/73.28 & 73.3/72.27 & 70.3/69.29 & 88.6/87.06 & 88.6/87.13 & 71.9/70.29 & 90.2/88.86 & 87.2/86.19 & 81.0/79.94 & 94.9/93.86 & 57.4/56.34 & 86.7/85.67 & 78.1/76.31 & 51.0/49.0 & 82.3/84.08 & 80.3/78.54 & 78.4/77.05 \\
SO-Net & 88.0/86.21 & 81.1/79.29 & 88.0/86.16 & 90.6/88.85 & 90.6/89.44 & 73.5/72.5 & 90.7/88.78 & 83.9/81.97 & 83.9/82.27 & 94.8/92.99 & 69.1/67.75 & 92.4/91.33 & 80.9/79.81 & 53.1/51.28 & 72.9/71.35 & 83.0/81.11 & 82.6/80.98 \\
PCNN & 85.5/84.32 & 79.5/78.1 & 79.5/77.78 & 73.2/72.0 & 90.9/89.55 & 73.2/71.3 & 90.3/88.9 & 90.0/88.04 & 86.9/85.45 & 95.8/93.9 & 69.0/67.04 & 92.1/90.64 & 86.3/84.5 & 51.0/49.88 & 83.1/81.55 & 83.3/85.14 & 82.5/80.96 \\
DGCNN & 86.7/85.21 & 77.8/75.93 & 86.7/88.35 & 77.8/76.6 & 90.6/88.74 & 74.7/72.98 & 91.2/89.49 & 87.5/86.32 & 86.8/85.01 & 95.7/94.35 & 66.3/64.95 & 81.1/82.85 & 81.1/79.63 & 63.5/62.17 & 74.5/75.76 & 82.6/81.36 & 82.0/80.91 \\
P2Sequence & 80.6/79.03 & 80.0/78.9 & 80.0/78.25 & 90.0/88.85 & 91.7/90.37 & 70.0/68.6 & 90.5/91.72 & 88.3/86.68 & 85.9/84.58 & 96.0/94.38 & 69.3/67.41 & 93.4/91.52 & 79.3/77.84 & 58.1/56.71 & 73.1/71.25 & 82.0/80.37 & 82.0/80.44 \\
PointCNN & 86.0/84.77 & 80.0/78.37 & 86.0/84.39 & 80.8/79.67 & 91.6/90.07 & 79.7/77.94 & 92.3/90.48 & 88.4/87.04 & 86.3/84.4 & 96.1/94.44 & 77.2/75.43 & 95.2/93.54 & 84.2/82.83 & 63.0/61.37 & 75.8/74.66 & 81.0/79.72 & 84.4/82.86 \\
PointASNL & 87.9/86.57 & 79.7/78.19 & 79.7/77.74 & 92.2/90.81 & 92.2/90.85 & 73.7/72.38 & 91.0/89.84 & 84.2/82.87 & 84.2/82.5 & 95.8/94.64 & 74.4/72.5 & 95.2/93.65 & 81.0/79.56 & 63.0/61.94 & 73.3/72.01 & 83.2/82.01 & 83.5/81.97 \\
RS-CNN & 84.0/85.58 & 86.2/88.02 & 81.2/79.68 & 81.1/79.4 & 91.2/89.91 & 81.1/79.73 & 91.2/89.71 & 91.1/89.99 & 86.7/85.27 & 96.3/97.65 & 73.7/72.39 & 94.7/92.99 & 83.5/81.86 & 60.5/59.2 & 77.7/76.38 & 83.2/81.38 & 84.0/82.72 \\
CurveNet & 81.9/83.61 & 75.2/73.9 & 74.9/72.98 & 74.9/73.7 & 93.0/91.89 & 74.9/73.21 & 93.0/91.48 & 86.1/84.96 & 84.7/83.69 & 95.6/96.85 & 66.7/67.86 & 92.7/93.99 & 81.6/80.18 & 63.4/62.01 & 82.3/80.82 & 82.1/80.75 & 81.1/80.22 \\
SPLATNet & 89.1/87.27 & 80.3/79.07 & 80.3/78.66 & 80.3/78.38 & 93.9/95.0 & 75.5/77.0 & 92.1/90.96 & 83.4/82.39 & 83.4/84.65 & 96.3/94.46 & 75.6/74.21 & 93.6/91.9 & 83.4/81.92 & 63.4/61.67 & 75.5/73.69 & 81.8/80.35 & 83.3/81.99 \\
SpiderCNN & 84.7/82.83 & 77.2/75.23 & 81.7/80.56 & 84.7/83.04 & 91.2/89.38 & 73.3/71.8 & 90.3/88.91 & 86.3/85.17 & 83.3/85.23 & 95.8/97.62 & 70.3/69.18 & 94.3/92.73 & 80.3/78.55 & 59.7/58.56 & 73.9/71.99 & 81.3/79.84 & 82.4/81.06 \\
KPConv & 87.2/85.93 & 81.1/79.77 & 81.1/79.86 & 81.1/83.09 & 91.1/89.34 & 76.2/74.92 & 88.4/87.14 & 88.4/87.01 & 82.0/80.45 & 96.2/94.34 & 78.1/76.15 & 95.9/97.7 & 85.4/83.97 & 69.2/67.6 & 80.2/78.69 & 83.0/81.26 & 84.0/82.58 \\
PA-DGC & 90.4/88.5 & 79.7/78.01 & 90.6/88.76 & 80.8/79.22 & 90.8/89.09 & 73.9/72.72 & 92.0/90.17 & 88.7/90.59 & 82.9/81.64 & 95.9/94.31 & 73.9/72.86 & 94.7/93.0 & 84.7/83.15 & 65.9/64.64 & 81.4/79.47 & 84.0/82.48 & 85.3/83.77 \\
PointMLP & 87.5/85.65 & 80.54/82.2 & 90.3/88.65 & 78.2/76.63 & 90.3/88.93 & 78.2/76.85 & 92.0/90.91 & 88.1/86.98 & 82.6/81.23 & 96.2/94.84 & 77.5/76.35 & 95.8/94.69 & 85.4/83.68 & 64.6/63.31 & 83.3/82.21 & 84.3/83.0 & 85.0/83.64 \\
\bottomrule
\end{tabular}
}
\vspace{-2mm}
\caption{Segmentation results on the ShapeNetPart dataset \cite{chang2015shapenet}.}
\vspace{-7mm}
\label{segment}
\end{table*}

\begin{table*}[t]
\centering
\resizebox{1\textwidth}{!}{
\scriptsize
\setlength{\tabcolsep}{3pt}
\begin{tabular}{lccccccc}
\toprule
 \textbf{Method} & \textbf{PointNet \cite{qi2017pointnet}} & \textbf{PointNet++ \cite{qi2017pointnet++}} & \textbf{DGCNN \cite{wang2019dynamic}} & \textbf{PointCNN \cite{li2018pointcnn}} & \textbf{GBNet \cite{qiu2022geometric}} & \textbf{Point Trans \cite{zhao2021point}} & \textbf{PointMLP \cite{marethinking}} \\
\midrule 
  Clean data & 0.682  & 0.779 & 0.781  & 0.785 & 0.805 & 0.828  & 0.854\\
 optical chaotic\cite{liu2023privacy} $\dagger$ & 0.041/0.081 & 0.048/0.084 & 0.052/0.087 & 0.056/0.089  & 0.062/0.092 & 0.070/0.096  & 0.079/0.099  \\

hybrid key \cite{li20243d} $\dagger$& 0.214/0.272 & 0.258/0.284 & 0.271/0.296 & 0.286/0.307 & 0.294/ 0.321& 0.306/0.336 & 0.314/0.349 \\
UMT \cite{wang2024unlearnable} (Class-wise test set)& 0.421/0.852 & 0.438/0.871 & 0.447/0.884 & 0.455/0.893  & 0.466/0.907 & 0.481/ 0.924& 0.493/0.939\\
UMT \cite{wang2024unlearnable} (  Permuted-wise test set)& 0.421/0.214  & 0.438/0.232  & 0.447/0.246 & 0.455/0.257 & 0.466/0.281 & 0.481/0.307 & 0.493/0.329\\

   Ours  & 0.071 / 0.664 & 0.078 / 0.759 & 0.084 / 0.771 & 0.089 / 0.775 & 0.098 /0.785 & 0.109/ 0.808 & 0.119/ 0.840 \\
\bottomrule
\end{tabular}
}

\caption{Comparison of different encryption methods  in the original and encrypted domains on the ScanObjectNN \cite{uy2019revisiting}. For each entry, the value before ``/'' denotes the classification accuracy when the classifier is trained on the original dataset and tested on the encrypted dataset; the value after ``/'' denotes the classification accuracy when the classifier is both trained and tested within the same protected/encrypted domain. $\dagger$ indicates results implemented by ourselves.}
\vspace{-6mm}
\label{tab:comp}
\end{table*}

\subsection{Downstream Task Evaluation}
\textbf{Classification Tasks: }
Table \ref{tab:scanobject-modelnet} compares the performance of various point cloud classification models on semantically encrypted versions of the {ScanObjectNN} \cite{uy2019revisiting} and ModelNet40 datasets \cite{wu20153d}. The results show that stronger classifiers are generally more capable of recognizing the target categories on encrypted data while avoiding misclassification back to the original categories, and therefore tend to achieve higher NCRR and lower OCLR. Furthermore, OLRR is consistently higher than OCLR, indicating that adaptive retraining attacks are indeed stronger than fixed classifiers; however, its value still remains at a relatively low level, suggesting that the original class information in semantically encrypted point clouds is still difficult to recover effectively. This trend is consistent across different datasets and backbone models, further demonstrating the stability and interpretability of our metrics under cross-backbone and cross-dataset settings.
As shown in Fig.\ref{transfer}, each triplet from left to right presents a target data, a source data, and the result obtained by applying the transfer table. Although a few residual features of the source class are still visible locally (e.g., certain edge lines or small parts not suppressed), the  point cloud structure is highly consistent with the target class: the body proportions, the relative positions of key components, and the geometric contours align with the target. 


\textbf{The Segmentation Tasks.}
Table \ref{segment} presents the segmentation results across multiple categories on the ShapeNetPart dataset \cite{chang2015shapenet} using our class-transferred (encrypted) point clouds.
Overall, the results show that our privacy-preserving transformation maintains high segmentation performance, with only a slight drop in accuracy compared to the original point clouds.
This verifies that the proposed method effectively preserves the usability of point cloud data while achieving privacy protection.
For categories with relatively fewer samples, a slightly larger performance fluctuation is observed, which can be attributed to the uneven distribution of shape priors in the latent space.

\begin{table*}[t]
\centering
\scriptsize
\resizebox{1\textwidth}{!}{
\setlength{\tabcolsep}{3pt}
\renewcommand{\arraystretch}{1.15}
\begin{tabular}{l|cccccccccccc|cccccccccccc}
\hline
\textbf{Setting}
& \multicolumn{12}{c|}{\textbf{ScanObjectNN}}
& \multicolumn{12}{c}{\textbf{ModelNet40}} \\
& \multicolumn{4}{c}{\textbf{PointNet}}
& \multicolumn{4}{c}{\textbf{PointNet++}}
& \multicolumn{4}{c|}{\textbf{PointMLP}}
& \multicolumn{4}{c}{\textbf{PointNet}}
& \multicolumn{4}{c}{\textbf{PointNet++}}
& \multicolumn{4}{c}{\textbf{PointMLP}} \\
\hline
& A & N & O & R & A & N & O & R & A & N & O & R
& A & N & O & R & A & N & O & R & A & N & O & R \\

1024  & 66.41 & 0.651 & 0.172 & 0.301 & 75.95 & 0.746 & 0.120 & 0.204 & 84.02 & 0.820 & 0.080 & 0.116
      & 88.10 & 0.850 & 0.090 & 0.162 & 90.30 & 0.860 & 0.085 & 0.145 & 92.90 & 0.890 & 0.065 & 0.094 \\
768   & 65.20 & 0.631 & 0.184 & 0.321 & 74.80 & 0.728 & 0.132 & 0.223 & 83.10 & 0.805 & 0.089 & 0.129
      & 87.30 & 0.839 & 0.098 & 0.173 & 89.60 & 0.850 & 0.092 & 0.156 & 92.10 & 0.881 & 0.071 & 0.103 \\
512   & 63.40 & 0.602 & 0.201 & 0.349 & 72.90 & 0.701 & 0.149 & 0.247 & 81.20 & 0.781 & 0.103 & 0.149
      & 85.90 & 0.821 & 0.111 & 0.190 & 88.20 & 0.836 & 0.103 & 0.171 & 90.90 & 0.868 & 0.081 & 0.118 \\
\hline
clean  & 66.41 & 0.651 & 0.172 & 0.301 & 75.95 & 0.746 & 0.120 & 0.204 & 84.02 & 0.820 & 0.080 & 0.116
       & 88.10 & 0.850 & 0.090 & 0.162 & 90.30 & 0.860 & 0.085 & 0.145 & 92.90 & 0.890 & 0.065 & 0.094 \\
0.005  & 65.90 & 0.642 & 0.178 & 0.311 & 75.40 & 0.738 & 0.126 & 0.214 & 83.70 & 0.813 & 0.085 & 0.123
       & 87.80 & 0.844 & 0.094 & 0.168 & 89.90 & 0.855 & 0.089 & 0.151 & 92.50 & 0.885 & 0.069 & 0.100 \\
0.01   & 64.90 & 0.625 & 0.189 & 0.327 & 74.30 & 0.721 & 0.137 & 0.229 & 82.90 & 0.798 & 0.093 & 0.134
       & 87.00 & 0.833 & 0.102 & 0.178 & 89.20 & 0.846 & 0.095 & 0.160 & 91.80 & 0.876 & 0.075 & 0.108 \\
0.02   & 63.10 & 0.598 & 0.206 & 0.356 & 72.60 & 0.694 & 0.153 & 0.251 & 81.00 & 0.774 & 0.107 & 0.153
       & 85.80 & 0.816 & 0.114 & 0.194 & 88.00 & 0.832 & 0.106 & 0.175 & 90.60 & 0.863 & 0.084 & 0.121 \\
\hline
\end{tabular}}
\vspace{-2mm}
\caption{Robustness analysis under varying point densities and noise levels. A: Acc, N: NCRR, O: OCLR, R: OLRR.}
\vspace{-6mm}
\label{tab:robustness_density_noise}
\end{table*}

\subsection{Comparison with  Point Cloud Encryption Methods}
Table \ref{tab:comp} compares the classification performance of different encryption methods under two tasks. Task 1 corresponds to training on the original data and testing on the encrypted data, and is used to evaluate the encryption effectiveness of each method. Task 2 corresponds to training and testing on the encrypted data, and is used to examine whether the encrypted point clouds still retain learnability. If only Task 1 were considered, geometry-destructive methods would naturally have an advantage, since their objective is precisely to reduce the recognition rate of the original categories as much as possible. However, this alone cannot answer another equally important question: whether the encrypted point cloud can still serve as a protected representation for subsequent downstream tasks. Therefore, we further introduce Task 2 to evaluate the downstream usability of the encrypted data.
To enable a fairer comparison with existing geometric point cloud encryption methods, we further construct a reversible destructive variant of our framework for Task 1, while the standard model is used for Task 2. This variant no longer treats ``stable transfer to a designated target category'' as the primary optimization objective. Instead, it enhances the destruction of the original semantics by weakening the target-category alignment constraint, strengthening the original-category suppression term, and moderately relaxing the smoothness and structure-preservation constraints. Meanwhile, we still retain the reversible backbone of the framework, so this variant remains a reversible encryption mode.
From the experimental results, geometry-based encryption methods, such as optical chaotic and hybrid key, usually achieve lower classification accuracy under Task 1, indicating that they are more advantageous in point cloud encryption. However, these methods also obtain low accuracy under Task 2, suggesting that their encrypted outputs are closer to unusable perturbation-like representations and are difficult to support for subsequent downstream processing. In contrast, although our method may not always outperform the strongest geometry-destructive encryption methods under purely destructive metrics, its reversible destructive variant can likewise significantly reduce the recognizability of the original categories in Task 1, while the standard model maintains high classification performance in Task 2. This indicates that the output of our method is not a semantically meaningless point cloud, but rather a protected representation that still preserves stable organizational structure and learnability.
For UMT, although it achieves relatively high accuracy in classification after encryption, its performance depends strongly on the correspondence between transformations and labels. Once this correspondence is disrupted at test time, its accuracy drops noticeably. This suggests that the high accuracy of UMT stems more from learning a fixed correspondence between transformation patterns and labels, rather than from modeling a stable semantic representation itself. In contrast, our method explicitly models inter-class relationships and a controllable transfer process.
Overall, this comparative experiment shows that the advantage of semantic-level methods over geometry-level methods does not lie in pursuing the strongest possible destructive encryption alone, but rather in achieving a more reasonable balance between suppressing the original semantics and preserving usability in the protected domain. In this way, the encrypted point cloud is not only protected, but can also continue to serve as a protected representation for downstream systems.

\subsection{Robustness under Point Density and Noise Variations}

Table \ref{tab:robustness_density_noise} presents the robustness analysis under different point cloud densities and noise levels using different classifiers on ScanObjectNN and ModelNet40. Overall, as the point cloud density decreases from 1024 to 768 and 512, or the input noise increases from clean to $\sigma=0.005, 0.01, 0.02$, metrics consistently decline. This indicates that when the geometric information becomes sparser or more corrupted, the transformed target semantics become less distinguishable, and the original semantics are more likely to remain or be recovered. Nevertheless, the degradation is generally gradual rather than abrupt, suggesting that the proposed method maintains good stability under moderate density reduction and noise perturbation.
Comparing different classifier backbones, PointMLP consistently achieves better results across both datasets and perturbation settings, indicating that stronger classifiers can better preserve target-category discriminability while suppressing original semantic leakage under degraded input conditions. In contrast, PointNet is more sensitive to point cloud sparsification and noise corruption, showing the most noticeable performance drop, while PointNet++ lies in between. This consistent trend across both datasets suggests that the proposed method preserves a stable privacy–utility trade-off across different classifier backbones.
From the dataset perspective, the performance degradation on ScanObjectNN is generally larger than that on ModelNet40, which is consistent with the fact that real-world scanned data are more complex. Even so, the variations of metrics on the more challenging ScanObjectNN remain within a reasonable range, without severe fluctuations, further demonstrating the robustness of the proposed method under different point cloud quality conditions.

\subsection{Incremental Adaptation to Unseen Classes}

\begin{table}[t]
\centering
\scriptsize
\resizebox{0.47\textwidth}{!}{
\setlength{\tabcolsep}{3pt}
\renewcommand{\arraystretch}{1.15}
\begin{tabular}{lcccccccccccc}
\hline
\textbf{Setting}
& \multicolumn{4}{c}{\textbf{box}}
& \multicolumn{4}{c}{\textbf{display}}
& \multicolumn{4}{c}{\textbf{door}} \\
\hline
& A & N & O & R
& A & N & O & R
& A & N & O & R \\
Full Joint Training
& 75.8 & 0.742 & 0.123 & 0.208
& 74.9 & 0.731 & 0.129 & 0.217
& 76.2 & 0.748 & 0.118 & 0.201 \\
Incremental Adaptation
& 74.2 & 0.721 & 0.136 & 0.226
& 73.1 & 0.708 & 0.145 & 0.239
& 74.8 & 0.728 & 0.132 & 0.221 \\
\hline
\end{tabular}}
\caption{Incremental adaptation results on unseen classes of ScanObjectNN using PointNet++. We treat \textit{box}, \textit{display}, and \textit{door} as unseen categories. A: Acc, N: NCRR, O: OCLR, R: OLRR.}
\vspace{-10mm}
\label{tab:unseen_scanobject}
\end{table}

Table \ref{tab:unseen_scanobject} reports the experimental results on unseen classes in ScanObjectNN, where \textit{box}, \textit{display}, and \textit{door} are treated as unseen classes, and PointNet++ is adopted as the classifier. Here, \emph{Full Joint Training} means that the unseen classes are included in the global training together with the other classes from the very beginning. In contrast, \emph{Incremental Adaptation} means that after the shared backbone has been fully trained, only the modules related to the new classes are locally adapted, without retraining the entire global framework. Full Joint Training achieves the best performance, because under this setting the unseen classes participate in the global joint optimization of the shared latent representation, inter-class energy relationships, and ODE-conditioned transitions from the start of training, and can therefore be regarded as the upper-bound performance.
By comparison, the results of Incremental Adaptation are slightly lower than those of Full Joint Training, but the gap between them remains generally small. This indicates that even if the unseen classes are not involved in the initial closed-set global training, competitive performance can still be achieved by locally adapting only the class-specific modules while keeping the shared backbone and shared latent space fixed. The reason is that, in our method, the components that truly depend on category differences are mainly the class embeddings, conditional modulation parameters, and the energy relationships between the new classes and the existing ones, whereas the Flow backbone, shared latent representation, and the main ODE dynamics learn generic structural representations in a category-shared space. Therefore, when introducing new classes, it is unnecessary to re-optimize the entire global framework; instead, effective integration can be achieved by only supplementing and updating these class-related components.
Overall, this experiment shows that although the primary setting of the current framework is closed-set category mapping, its backbone, latent representation, and main ODE dynamics are all shared structures. As a result, new classes can be  incorporated through local incremental adaptation, without requiring full retraining of the entire framework.

\begin{table}[t]
\centering
\resizebox{0.47\textwidth}{!}{
\begin{tabular}{lcccc}
\toprule
Method & Params (M) & NCRR $\uparrow$ & OCLR $\downarrow$ & Acc $\uparrow$ \\
\midrule
15$\times$  Flow        & 159.75 & 0.71 & 0.18 & 0.8554 \\
Shared + LoRA            & 20.51  & 0.75 & 0.12 & 0.8769 \\
Shared + LoRA + FiLM & 21.12  & 0.81 & 0.08 & 0.9032 \\
\bottomrule
\end{tabular}
}
\caption{Comparison of parameter counts and privacy/utility metrics for different Flow models on the {ModelNet40} datasets \cite{wu20153d}.}
\vspace{-10mm}
\label{tab:ncrr-oclr-params}
\end{table}

\subsection{Flow Parameter Efficiency}
Table~\ref{tab:ncrr-oclr-params} compares per-class independent flow models with the proposed shared-backbone flow augmented by LoRA and FiLM, in terms of parameter size as well as privacy and utility metrics on the ScanObjectNN dataset~\cite{uy2019revisiting}.
Here, Acc denotes the classification accuracy obtained by retraining a classifier on the transformed point clouds.
Compared to the baseline that trains a separate flow model for each class, introducing a shared backbone with LoRA reduces the total number of parameters from 159.75M to 20.51M, corresponding to an 87\% reduction, while simultaneously improving all three evaluation metrics.
This result indicates that even with a substantially compressed model, both new-class recognizability and privacy protection can be enhanced.
Further incorporating FiLM modulation increases the parameter count by only approximately 3\%, yet yields the best overall performance.


\subsection{Analysis of Class Transfer Table}

\begin{figure}[t]
\begin{center}
\includegraphics[width=6cm, height=6cm]{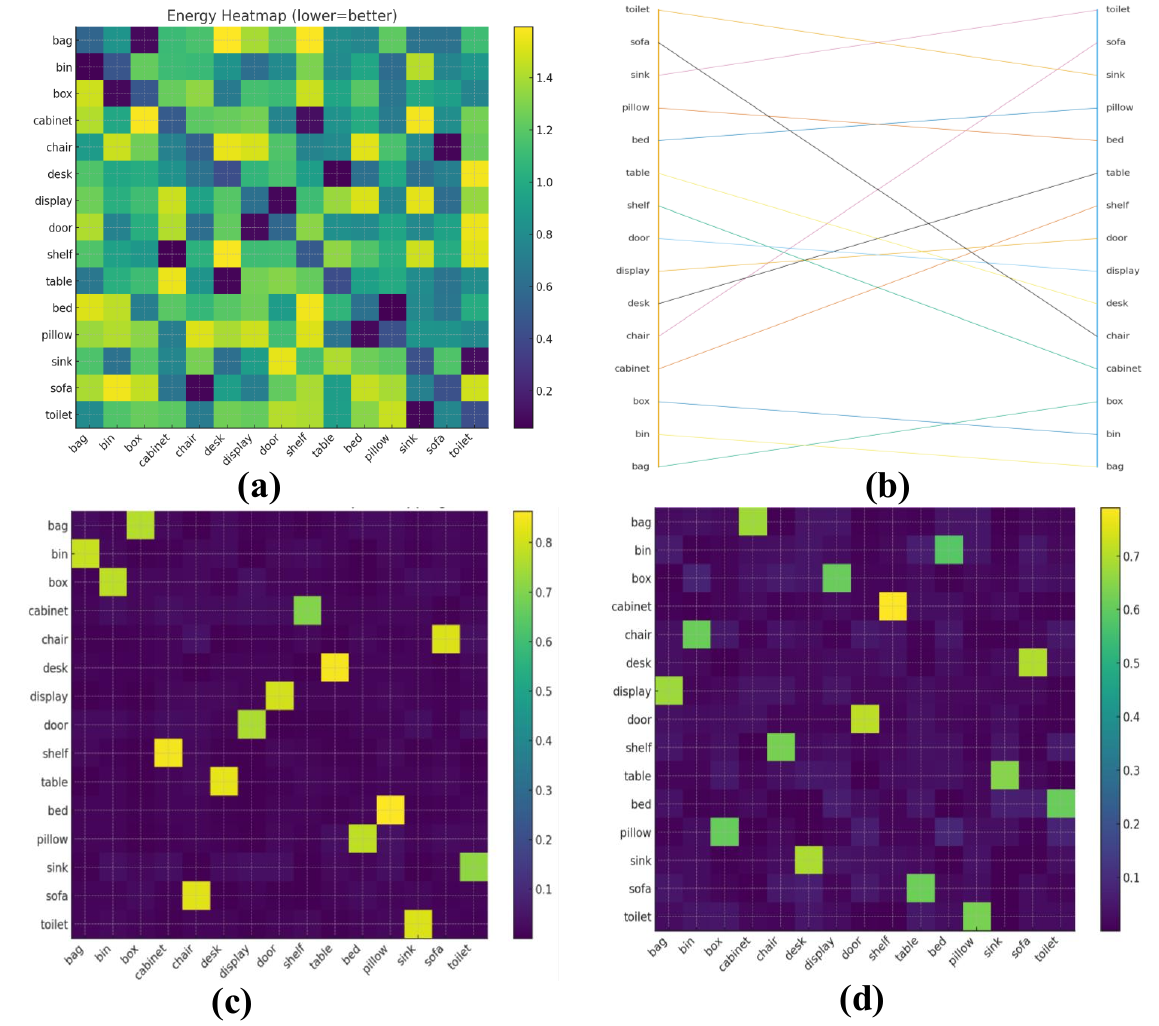}
\end{center}
\vspace{-3mm}
 \caption{Energy and effectiveness of the class-transfer table on the ScanObjectNN dataset.
(a) Energy matrix heatmap: each entry \(E_{ij}\) denotes the conversion energy from source class \(i\) to target class \(j\) (lighter color = lower energy).
(b) Source\(\rightarrow\)target pairs obtained by energy-based one-to-one matching; thicker lines indicate lower energy (better pairs).
(c) Confusion matrix after applying the transfer table.
(d) Confusion matrix under random mapping.}
\label{fig:scanobjectnn-energy-mapping}
\vspace{-3mm}
\end{figure}

Fig.~\ref{fig:scanobjectnn-energy-mapping} illustrates the effectiveness of the proposed class-transfer table on the ScanObjectNN dataset.
We define an energy function $E_{ij}$ to quantify the conversion cost from a source class $i$ to a target class $j$, and derive a one-to-one class-transfer table through global minimum-cost matching.
As shown in the energy heatmap, semantically and geometrically related class pairs (e.g., \{chair\}$\leftrightarrow$\{sofa\}, \{sink\}$\leftrightarrow$\{toilet\}) consistently exhibit low-energy paths, which align well with the thick-line correspondences in the resulting mapping diagram.
Applying this class-transfer table for semantic transformation leads to a clear structural change in the confusion matrix.
Specifically, each source-class row displays a pronounced peak at the designated target-class column, while responses at the original-class column are strongly suppressed.
This behavior directly corresponds to an increase in NCRR and a decrease in OCLR.
By contrast, random class mappings lack energy-based constraints, resulting in more diffuse prediction distributions, reduced aggregation toward target classes, and consequently lower NCRR and higher OCLR.
Overall, the energy-driven global matching strategy enables stable and consistent ``source$\rightarrow$target'' aggregation while inducing only minimal latent-space deformation.
In comparison, random transfer strategies fail to reliably guide samples toward target classes and are more prone to reverting to original-class predictions, underscoring the effectiveness of the proposed energy-based class-transfer table.

\begin{figure}[t]
\begin{center}
\includegraphics[width=8cm, height=5cm]{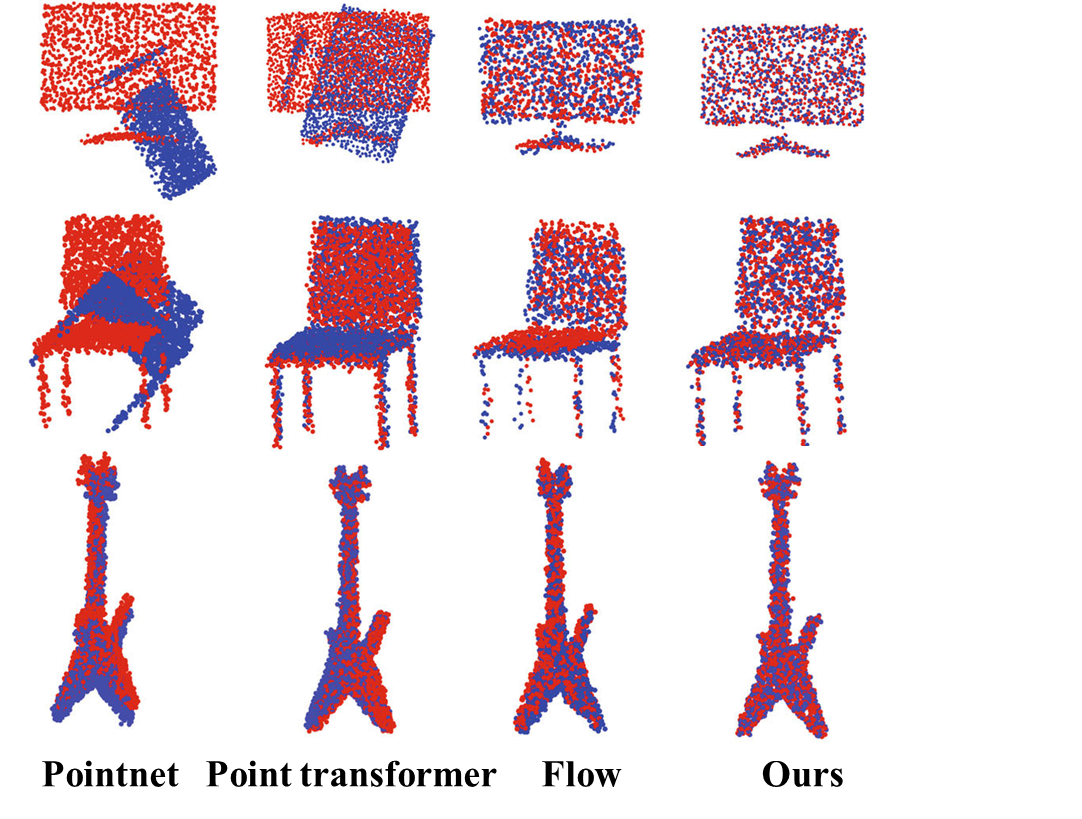}
\end{center}
\vspace{-3mm}
\caption{Qualitative comparison of encryption--decryption reconstruction fidelity across different backbones. 
Red points denote the original point clouds before encryption, while blue points denote the decrypted point clouds.}
\label{fig:backbone-encrypt-decrypt}
\vspace{-2mm}
\end{figure}
\subsection{Encryption Reconstruction}
As shown in Fig. \ref{fig:backbone-encrypt-decrypt}, we compare different models on three representative objects to evaluate the geometric fidelity after the “encryption → decryption” process. Red points denote the original shapes before encryption, while blue points indicate the decrypted results; ideally, the two should overlap almost perfectly. The PointNet-based scheme, which relies heavily on global pooling, fails to capture fine-grained local geometry, leading to noticeable misalignment and collapse artifacts between the original and recovered shapes (the overlap between red and blue points is clearly limited). Point Transformer recovers the overall contour more accurately, but still exhibits deviations on object boundaries and thin structures (e.g., chair legs and the guitar neck). The vanilla Flow model is able to reproduce most of the global geometry, yet minor local deformations and noise remain visible. In contrast, ours method yields decrypted point clouds that almost perfectly coincide with the originals; red and blue points are visually indistinguishable, indicating that the learned class-conditional invertible mapping achieves much higher geometric fidelity in both global structure and local details.

\subsection{Analysis of LoRA and FiLM}

\textbf{LoRA:} 
Table~\ref{tab:lora-film-ablation} reports the results of using different LoRA ranks under the shared Flow backbone. We observe that when increasing the rank from $r=4$ to $r=8$, the NCRR improves from 0.74 to 0.75, the OCLR decreases from 0.13 to 0.12, and the Acc rises from 0.8720 to 0.8769, while the number of parameters grows from 15.58M to 20.51M. Overall, this configuration leads to a stable performance gain. However, when further increasing the rank to $r=16$, although results still show slight improvements, the marginal benefits become very limited, whereas the parameter count increases further to 30.37M. Taken together, a medium-sized LoRA configuration ($r=8$) achieves a better trade-off between parameter efficiency and the privacy–utility metrics, indicating that LoRA is highly parameter-efficient for this task and that overly large ranks are unnecessary.

\textbf{FiLM:} 
Table~\ref{tab:lora-film-ablation} further investigates the impact of different FiLM insertion positions under the shared Flow + LoRA framework. Compared with the Shared+LoRA baseline without any FiLM, inserting FiLM only in the latter half of the coupling blocks (last 10 blocks) increases the number of parameters by merely about 0.31M (from 20.51M to 20.82M), yet significantly boosts NCRR to 0.79, reduces OCLR to 0.09, and raises Acc to 0.8990. This indicates that channel-wise modulation applied to higher-level features is already sufficient to capture most of the semantic information. When FiLM is further applied to all coupling blocks, the parameter count slightly grows to 21.12M, while results are further and Acc reaches 0.9032. Overall, with limited parameter overhead, FiLM-based fine-grained modulation of the Flow latents makes a clear contribution to improving new-class recognizability and suppressing original-class leakage, and inserting FiLM only into a subset of critical layers is already able to capture the majority of these benefits.



\begin{table}[t]
\centering
\resizebox{0.47\textwidth}{!}{
\begin{tabular}{lcccc}
\toprule
Method & Params (M) & NCRR $\uparrow$ & OCLR $\downarrow$ & Acc $\uparrow$ \\
\midrule
Shared + LoRA ($r=4$)  & 15.58 & 0.74 & 0.13 & 0.8720 \\
Shared + LoRA ($r=8$)  & 20.51 & 0.75 & 0.12 & 0.8769 \\
Shared + LoRA ($r=16$) & 30.37 & 0.76 & 0.11 & 0.8785 \\
\midrule
Shared + LoRA (no FiLM)             & 20.51 & 0.75 & 0.12 & 0.8769 \\
Shared + LoRA + FiLM (last 10 blks) & 20.82 & 0.79 & 0.09 & 0.8990 \\
Shared + LoRA + FiLM (all blks)     & 21.12 & 0.80 & 0.08 & 0.9032 \\
\bottomrule
\end{tabular}
}
\caption{Ablation studies on LoRA rank and FiLM placement under the shared Flow setting.}
\label{tab:lora-film-ablation}
\vspace{-10mm}
\end{table}

\subsection{Energy Constraint in  ODEs}

\begin{figure}[t]
\begin{center}
\includegraphics[width=7.5cm, height=5cm]{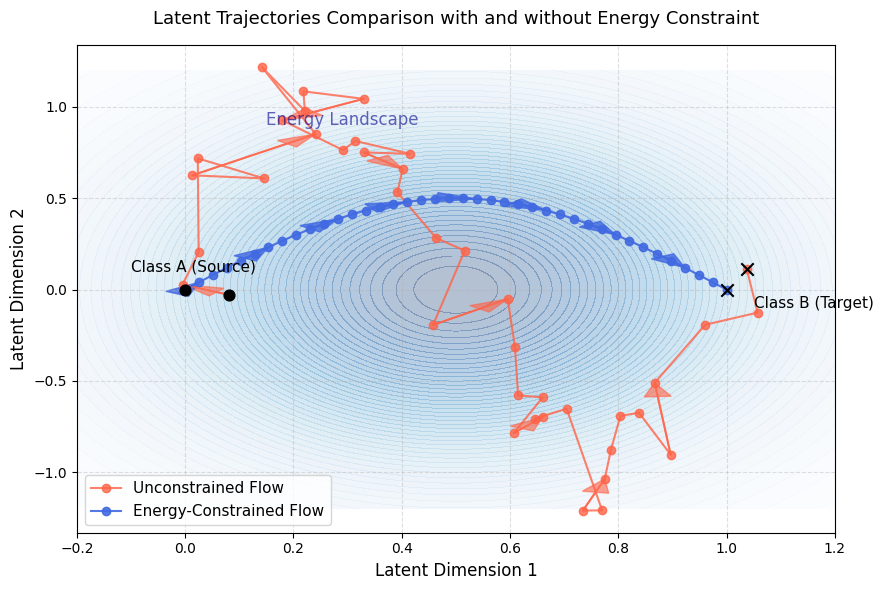}
\end{center}
\vspace{-7mm}
\caption{Comparison of latent trajectories with and without energy constraint.
The figure visualizes class transitions from Class A to B in latent space. The energy-constrained flow (blue curve) evolves smoothly along low-energy valleys, while the unconstrained flow (orange path) exhibits irregular and unstable motion across high-energy regions.}
\label{traject}
\vspace{-3mm}
\end{figure}

The visualization of latent trajectories in Fig \ref{traject} demonstrates the stabilizing effect of the proposed energy conservation constraint during class-wise transformation.
As shown in the figure, the unconstrained flow (red trajectory) exhibits oscillatory and irregular transitions in latent space, frequently crossing energy levels and resulting in non-smooth mappings.
In contrast, the energy-constrained flow (blue) follows a much smoother and shorter path between the source and target classes, maintaining consistent energy levels along the trajectory.
This indicates that the constraint effectively regularizes the neural ODE dynamics, preventing gradient explosion and ensuring that the transformation evolves along iso-energy surfaces, leading to stable and reversible mappings.

\subsection{Module Loss Analysis}
On the {ModelNet40} dataset \cite{wu20153d}, we perform a detailed analysis of each module.
\textbf{Diffusion Module:}
Table~\ref{tab:ablation_combined} 
presents the analysis of diffusion module.
First, without any diffusion alignment, the result shows that latent distribution of the transformed point clouds still exhibits a clear mismatch from that of the true target class. When we introduce the standard noise-reconstruction loss in latent space result indicates that denoising and smoothing in the latent space help enhance target-class recognizability while attenuating residual information from the original class. However, this denoising process does not explicitly enforce distribution alignment, so its benefits remain limited.
Building on the denoising loss, further adding an alignment term with respect to the Flow target distribution yields the best performance. The alignment loss drives the diffusion-generated latents closer to the true target-class manifold, thereby improving new-class recognition while further suppressing leakage of the original class. This module is used only during training as a latent-space distribution alignment regularizer; no diffusion steps are performed during encryption or decryption at inference time.
\begin{table}[t]
\centering
\resizebox{1\linewidth}{!}{
\begin{tabular}{llccc}
\toprule
Module & Setting & NCRR $\uparrow$ & OCLR $\downarrow$ & Acc $\uparrow$ \\
\midrule
& w/o Diffusion 
& 0.78 & 0.11 & 0.8910 \\
& Diffusion 
& 0.79 & 0.09 & 0.8985 \\
& Diffusion + $\mathcal{L}_{\text{align}}$ 
& \textbf{0.81} & \textbf{0.08} & \textbf{0.9032} \\
\midrule
& Scale 
& 0.77 & 0.11 & 0.8875 \\
& Scale + $\mathcal{L}_{\text{contrast}}$ 
& 0.79 & 0.09 & 0.8984 \\
& Scale + $\mathcal{L}_{\text{contrast}}$ + $\mathcal{L}_{\text{disp}}$
& \textbf{0.81} & \textbf{0.08} & \textbf{0.9032} \\
\bottomrule
\end{tabular}
}
\caption{Ablation studies on the diffusion module and dual-level scale control.}
\vspace{-10mm}
\label{tab:ablation_combined}
\end{table}
\textbf{Dual-Level Scale Control Losses:}
Table~\ref{tab:ablation_combined} presents the analysis of the Dual-Level Scale Control module.
We introduce only the scale normalization term (row 1). The result shows that the class-conditional latent space exhibits significant inter-class overlap.
Further adding the semantic contrastive loss $\mathcal{L}_{\text{contrast}}$ (row 2) leads to a continued
improvement. The contrastive loss pulls each
sample closer to its own class center while pushing it away from other class centers, so that the
class-conditional latent space exhibits a ``compact-within-class and separable-across-classes'' structure.
Finally, after incorporating the class embedding dispersion regularizer $\mathcal{L}_{\text{disp}}$ (row 3),
all metrics reach best result. The dispersion regularizer
enforces sufficient distances between different class embeddings, preventing them from collapsing to similar
directions and further strengthening inter-class separability.
\begin{table}[t]
\centering
\resizebox{0.47\textwidth}{!}{
\begin{tabular}{cccccc}
\toprule
$\mathcal{L}_{\text{align}}$ 
& $\mathcal{L}_{\text{rev}}$ 
& $\mathcal{L}_{\text{energy}}$
& NCRR $\uparrow$ & OCLR $\downarrow$ & Acc $\uparrow$ \\
\midrule
 & &  & 0.78 & 0.11 & 0.8920 \\
\checkmark &  &  & 0.79 & 0.10 & 0.8965 \\
\checkmark & \checkmark &  & 0.80 & 0.09 & 0.9008 \\
\checkmark & \checkmark & \checkmark & \textbf{0.81} & \textbf{0.08} & \textbf{0.9032} \\
\bottomrule
\end{tabular}
}
\caption{Ablation study on the ODE-related losses. 
distribution alignment loss $\mathcal{L}_{\text{align}}$, 
reversibility consistency loss $\mathcal{L}_{\text{rev}}$, 
and local energy smoothness loss $\mathcal{L}_{\text{energy}}$.
}
\vspace{-9mm}
\label{tab:ode-loss-ablation}
\end{table}
\textbf{Category Transition Flow.}
In Table~\ref{tab:ode-loss-ablation}, 
baseline (first row), these results indicate that relying solely on ODE fitting is insufficient to achieve satisfactory semantic encryption.
When we add $\mathcal{L}{\text{align}}$ (second row), the result shows the loss explicitly forces the ODE terminal distribution to approach the true target-class latent distribution, making the transformed samples more “similar” to the desired target category.
In the third row, we further add $\mathcal{L}{\text{rev}}$ by constraining the round-trip consistency from source to target and back to source. $\mathcal{L}{\text{rev}}$ encourages the ODE trajectories to be more reversible while preserving geometric structure, thereby reducing extreme paths and numerical instability.
Finally, after including $\mathcal{L}{\text{energy}}$ (last row), by penalizing sharp variations in the local energy field, $\mathcal{L}{\text{energy}}$ guides the ODE trajectories to traverse low-energy regions preferentially and avoid high-energy, high-risk areas, which further suppresses original-class leakage.

\begin{table}[t]
\centering
\resizebox{0.47\textwidth}{!}{
\begin{tabular}{lllllllll}
\multicolumn{1}{c}{Flow} & 
\multicolumn{1}{c}{Diffusion} & 
\multicolumn{1}{c}{Dual} &  
\multicolumn{1}{c}{Cate} & 
\multicolumn{1}{c}{Trans Flow} & 
\multicolumn{1}{c}{ NCRR $\uparrow$  } & 
\multicolumn{1}{c}{OCLR $\downarrow$ } & 
\multicolumn{1}{c}{Acc $\uparrow$} & 
\\ \hline 
  &  & &  &  & 0.71  & 0.18 & 0.8554,   
\\  \hline 
\checkmark    &  &  &  &  &0.74   & 0.14& 0.8695 
\\  \hline 
\checkmark   &\checkmark &   &  &  & 0.75  &0.12  & 0.8769
\\  \hline  
\checkmark   &\checkmark & \checkmark &  &  & 0.76  &0.12   &.8830. 
\\  \hline  
\checkmark   &\checkmark & \checkmark &\checkmark  &  & 0.78,  & 0.10 & 0.8905
\\  \hline  
\checkmark   &\checkmark & \checkmark &\checkmark  & \checkmark & \textbf{0.81} & \textbf{0.08} & \textbf{0.9032}
\\  \hline  
\end{tabular}}
\caption{Ablation study on the  {ModelNet40} dataset \cite{wu20153d}. Dual: DDual-Level Scale Control. Cate: Category Transition Table. Trans Flow: Category Transition Flow.}
\vspace{-9mm}
\label{tab:overall-ablation}
\end{table}

{
\subsection{Sensitivity Analysis of Loss}
\textbf{Diffusion Module: Loss Design and Sensitivity.}
As shown in Fig.~\ref{loss_weight}(a), we study the sensitivity of the proposed metrics with respect to the alignment loss weight $\lambda_{\text{align}}$. When $\lambda_{\text{align}}$ is very small (e.g., $0.01$--$0.02$), the alignment effect is weak, leading to degraded performance. As $\lambda_{\text{align}}$ increases to the range of $0.05$--$0.2$, the metrics improve and form a relatively stable plateau: NCRR rises to about $0.81$, OCLR decreases to $0.08$, and Acc reaches its peak of approximately $0.9032$ at $\lambda_{\text{align}} = 0.1$. When $\lambda_{\text{align}}$ is further increased ($0.4$--$1.0$), the performance gain becomes very limited, and the metrics even gradually decline, while $1-\text{OCLR}$ remains nearly flat, indicating mild over-regularization. Overall, these results suggest that our default setting lies in a relatively flat and near-optimal region, and the method is not highly sensitive to $\lambda_{\text{align}}$.
\textbf{Dual-Level Scale Control: Loss Design and Sensitivity.}
We further investigate the sensitivity of the Dual-Level Scale Control to the dispersion weight $\lambda_{\text{disp}}$. As shown in Fig.~\ref{loss_weight} (b), when $\lambda_{\text{disp}}$ is too small (e.g., $0.01$--$0.02$), the class embedding dispersion is under-regularized: NCRR stays around $0.79$--$0.802$, OCLR remains relatively high ($0.10$--$0.089$), and Acc is below $0.90$. Increasing $\lambda_{\text{disp}}$ to $0.05$ yields the best overall performance, indicating that a moderate dispersion strength is beneficial for pushing class embeddings apart and reducing original-class leakage without harming utility. Beyond this point, further enlarging $\lambda_{\text{disp}}$ ($0.1$--$1.0$) leads to only mild but consistent degradation, suggesting that overly strong dispersion may over-penalize class proximity and introduce unnecessary distortion to the latent structure. Overall, the curve exhibits a clear peak around $\lambda_{\text{disp}} = 0.05$, while still showing that the method is reasonably robust in the nearby range.
\begin{figure}[t]
\begin{center}
\includegraphics[width=7cm, height=5.5cm]{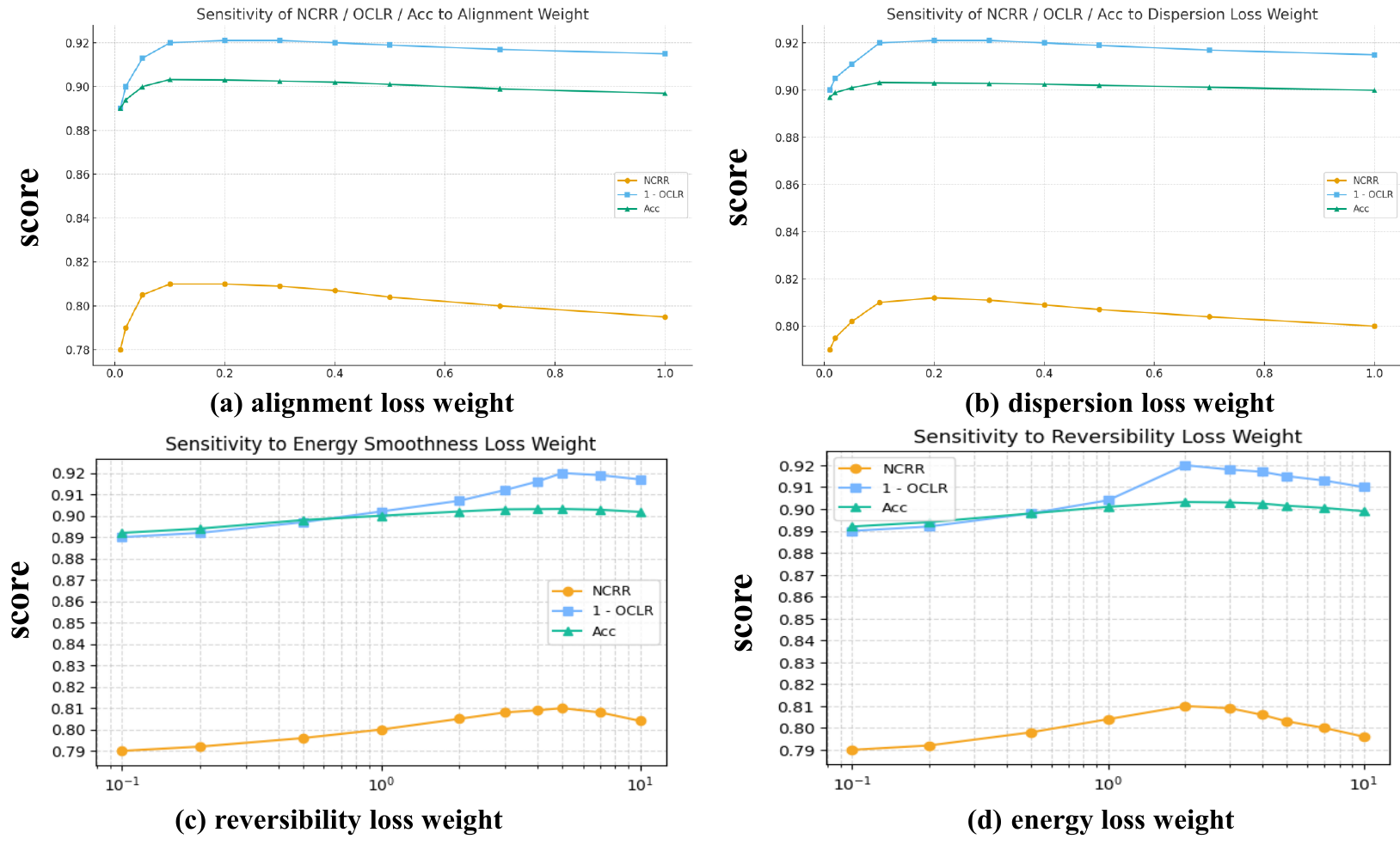}
\end{center}
\vspace{-6mm}
\caption{Sensitivity analysis of the loss weights.  For (a) and (b), we vary $\lambda_{\text{align}}$ and $\lambda_{\text{disp}}$ from 0.01 to 0.1. For (c) and (d), we vary $\lambda_{\text{rev}}$ and $\lambda_{\text{energy}}$ within $[0.1, 10]$, while keeping the other loss weights fixed.} 
\label{loss_weight}
\vspace{-3mm}
\end{figure}
\textbf{Category Transition Flow: Loss Design and Sensitivity.}
To further understand the behavior of the ODE-based category transition module, we conduct a weight sensitivity analysis on the reversibility loss and the energy smoothness loss. Fig.~\ref{loss_weight}(c) varies the reversibility weight $\lambda_{\text{rev}}$ while keeping the other loss weights fixed. When $\lambda_{\text{rev}}$ is too small (e.g., $0.1$--$0.5$), the reversibility constraint is insufficient and the metrics remain low, indicating that the learned trajectories are more likely to drift and cannot guarantee a stable round-trip mapping between source and target latents. As $\lambda_{\text{rev}}$ increases to a moderate range (around $\lambda_{\text{rev}} \approx 2$), all three metrics consistently improve and reach their peak, showing that an appropriate reversibility regularization helps the ODE preserve geometric structure and reduce original-class semantic leakage. Further increasing $\lambda_{\text{rev}}$ leads to a slight decline in the metrics, suggesting that overly strong constraints over-restrict the flow and weaken its ability to adapt to complex category transitions.
Fig.~\ref{loss_weight}(d) presents a similar analysis for the energy smoothness weight $\lambda_{\text{energy}}$ with $\lambda_{\text{rev}}$ fixed. When $\lambda_{\text{energy}}$ is very small, the ODE trajectories are not sufficiently guided by the energy field and may pass through high-energy, high-risk regions, resulting in poor metric values. As $\lambda_{\text{energy}}$ increases to a moderate value (around $\lambda_{\text{energy}} \approx 5$), all metrics reach their best levels. This indicates that energy regularization can effectively guide the trajectories along smoother, lower-energy paths, allowing them to better match the target-class manifold. However, when $\lambda_{\text{energy}}$ becomes too large, the excessive smoothness constraint starts to over-regularize the flow and slightly harms downstream performance. Overall, these two sensitivity curves verify that our chosen weights lie near a stable “sweet spot.”

\subsection{Sensitivity Analysis of Module}
In Fig \ref{module1}, We further investigate the weight sensitivity of the overall loss of the three modules---diff loss, dual loss, and ODE loss---by varying one weight at a time in the range $[0.01, 0.40]$ while keeping the other two fixed at their optimal values, in order to analyze their impact on the behavior of the encryption model.
\textbf{Diff loss weight.}
When the weight of the diff loss $\lambda_{\text{diff}}$ varies around its optimal value ($0.03$), the three metrics change only slightly and form a relatively broad plateau. This indicates that the model is quite robust to moderate perturbations of $\lambda_{\text{diff}}$: the diff term improves the structure of the encrypted representations, but as long as its weight is not too large, the influence on metrics remains mild. When $\lambda_{\text{diff}}$ becomes relatively large (e.g., $\ge 0.25$), all three metrics start to decrease, suggesting that over-emphasizing this term constrains the latent distribution too strongly and degrades both recognition and privacy.
\textbf{Dual loss weight.}
In contrast, the curves for the dual loss show much sharper sensitivity. As the weight $\lambda_{\text{dual}}$ deviates from its optimal value ($0.05$), all metrics drop more noticeably, and $1-\text{OCLR}$ decreases the fastest. This means the dual loss is the most critical term for controlling category leakage: if $\lambda_{\text{dual}}$ is too small, the transformation is not sufficiently regularized and the original class information leaks out; if it is too large, the mapping becomes over-constrained and hurts metrics. Therefore, $\lambda_{\text{dual}}$ must be tuned in a relatively narrow range.
\textbf{ODE loss weight.}
The ODE loss weight $\lambda_{\text{ode}}$ exhibits an intermediate behavior between the above two. Around the optimal value ($0.10$), all three metrics remain high and change smoothly, forming a wide but clearly peaked region. When $\lambda_{\text{ode}}$ is too small, the latent trajectories become less well-behaved and performance degrades; when it is too large, the dynamics are overly restricted, which also leads to a gradual drop in metrics. This shows that the ODE loss stabilizes the transformation, but its effective range is wider than that of the dual loss.

\begin{figure*}[t]
\begin{center}
\includegraphics[width=15cm, height=3cm]{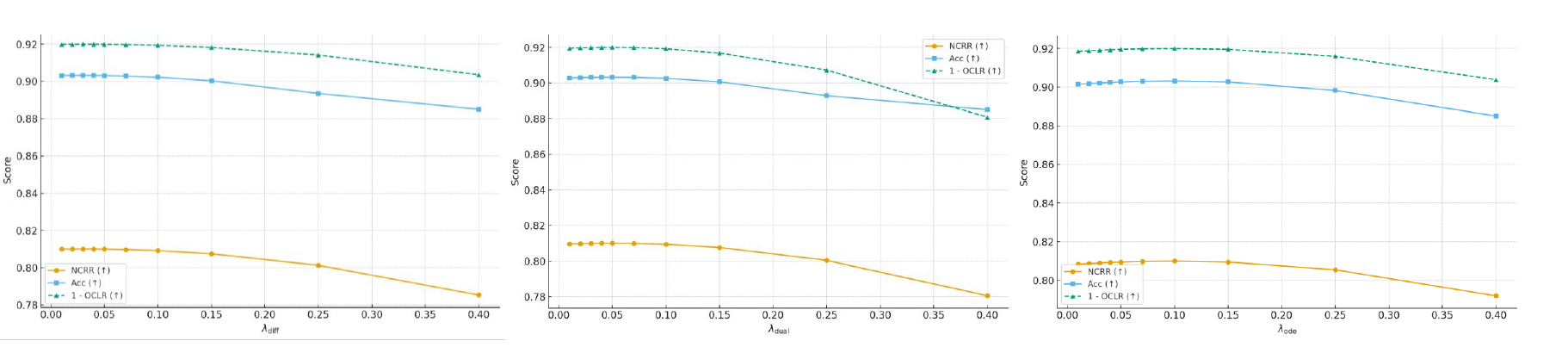}
\end{center}
\vspace{-7mm}
\caption{Sensitivity of the three module loss weights. 
We vary one weight within $[0.01, 0.40]$ while fixing the other two at their optimal values. 
The left figure corresponds to the diff module loss, the middle figure corresponds to the dual module loss, and the right figure corresponds to the ODE module loss. }
\label{module1}
\vspace{-5mm}
\end{figure*}
}

\subsection{Ablation Study}

We conduct ablation studies on the ModelNet40 dataset~\cite{wu20153d}.
Table~\ref{tab:overall-ablation} shows the stepwise ablation results.
The first row reports the performance of a conventional normalizing flow baseline, which exhibits limited effectiveness for semantic encryption of point clouds.
The second row corresponds to the shared-backbone Flow augmented with LoRA and FiLM.
Despite a substantial reduction in the number of trainable parameters, this configuration already alleviates inter-class entanglement, indicating that a shared backbone combined with lightweight conditional modulation provides a strong baseline.
In the third setting, the diffusion module is further introduced.
By denoising and smoothing the target latent space during training, this module leads to a more stable and better-aligned target latent distribution generated by the flow.
The fourth row incorporates the Dual-Level Scale Control mechanism, which calibrates geometric and semantic scale discrepancies across categories within a dimensionless latent space.
As a result, intra-class representations become more compact, while inter-class separation is further enhanced.
In the fifth configuration, we introduce the Category Transition Table.
The explicitly defined category mapping yields semantically meaningful and energy-efficient source--target pairs, encouraging transformed samples to aggregate toward the intended target categories.
Finally, the full model adds the Category Transition Flow.
Compared to linear interpolation, the Neural ODE learns a continuous latent flow field that enables adaptive and smooth transitions from source to target classes, bypassing high-energy regions and following more geometrically and semantically coherent trajectories.

\vspace{-1mm}
\section{Limitations}
\vspace{-1mm}
Although our model achieves promising results for semantic-level privacy protection of point clouds, several limitations still remain. First, the current framework mainly operates under a closed-set category setting. Although we further verify its incremental adaptation capability to unseen categories, the stability and scalability of the method under fully open-set scenarios, where a large number of new categories are continuously introduced, still require more systematic investigation. Second, our method is not designed to maximize the destruction of original-class recognizability; instead, it emphasizes the balance among suppression of original semantics, protected-domain learnability, and authorized recoverability. Therefore, under evaluation criteria that focus solely on extreme destructiveness, it may not always achieve the lowest original-class recognition rate. Finally, this work mainly focuses on category-level semantic privacy, while protection of instance-level identity, local attributes, or finer-grained sensitive scene information has not yet been explicitly modeled. This also represents an important direction for future extension.

\vspace{-2mm}
\section{Conclusion}
\vspace{-2mm}
In this paper, we presented a semantic encryption framework for 3D point clouds that enables privacy protection through category-conditioned reversible transformation. Our framework reduces the risk of original-class semantic leakage while preserving the usability of protected representations for downstream tasks and allowing authorized recovery.
Through a shared flow-based latent space, Category Transition Energy, and Neural ODE-based reversible dynamics, our model  achieves semantic transfer between classes and provides controllable protection in the semantic domain. 
Extensive experiments demonstrate that the proposed method effectively suppresses original semantic leakage while maintaining strong downstream learnability for tasks such as classification and segmentation. These results suggest that semantic encryption offers a promising alternative to destructive perturbation-based protection. We provides a new paradigm for making shared 3D point cloud simultaneously private, reusable, and recoverable, and opens a practical direction for trustworthy 3D data sharing.

\vspace{-2mm}
\section{Data Availability Statement}
\vspace{-2mm}
This study uses two types of publicly available point cloud benchmark datasets: point cloud classification datasets and point cloud semantic/part segmentation datasets. The specific dataset names, versions, and citations are provided in the Experimental Setup and Datasets section. No new raw point cloud data were collected in this work, and all original data were obtained from official public sources and used in accordance with their respective licenses.
This study does not involve any personally identifiable information. Its privacy risk mainly lies in the potential leakage of point cloud semantic class information, which is systematically evaluated in the paper using relevant privacy-related metrics.

\vspace{-2mm}
\section*{Acknowledgement}
\vspace{-2mm}
This work is supported by NSF Grants NO. 2340882, 2334624, 2334246, and 2334690. 
\bibliography{sn-bibliography}

\end{document}